\documentclass[accepted]{uai2026} % for initial submission
\usepackage[american]{babel}
\usepackage{microtype}
\usepackage{graphicx}
\usepackage{subcaption}
\usepackage{booktabs} % for professional tables
\usepackage{bm}
\usepackage{hyperref}

\usepackage[ruled,vlined,linesnumbered]{algorithm2e}
\SetCommentSty{footnotesize}

\newcommand{\F}{\mathcal{F}}

\usepackage{mathrsfs}
\newcommand{\PM}{\mathscr{P}}
\newcommand{\CM}{\mathscr{C}}

\usepackage{amsmath}
\usepackage{amssymb}
\usepackage{mathtools}
\usepackage{amsthm}
\usepackage{graphicx}
\usepackage{amsmath}
\usepackage{amsfonts}

\usepackage[table]{xcolor}
\usepackage{adjustbox}
\usepackage{booktabs}
\usepackage{cuted}
\usepackage{float}
\newfloat{eqfloat}{tbp}{loe}
\floatname{eqfloat}{Equation Block}

\usepackage[capitalize,noabbrev]{cleveref}

\theoremstyle{plain}

\theoremstyle{definition}

\theoremstyle{remark}

\usepackage[textsize=tiny]{todonotes}

\usepackage{natbib} % has a nice set of citation styles and commands
\usepackage{mathtools} % amsmath with fixes and additions
\usepackage{booktabs} % commands to create good-looking tables
\usepackage{tikz} % nice language for creating drawings and diagrams
\usepackage{float}

\usepackage[table, dvipsnames]{xcolor}

\hypersetup{
    citecolor=CadetBlue,
    colorlinks=true,
    linkcolor=CadetBlue,
    filecolor=magenta,      
    urlcolor=cyan}

\colorlet{table_baselines}{CadetBlue!10}

\title{Prior-Fitted Functional Flow:\\ In-Context Generative Models for Pharmacokinetics}

\title{Prior-Fitted Functional Flow:\\
In-Context Generative Models for Pharmacokinetics}

\author{
\textbf{César Ojeda}$^{1}$ \quad
\textbf{Niklas Hartung}$^{1}$ \quad
\textbf{Wilhelm Huisinga}$^{1}$ \quad
\textbf{Tim Jahn}$^{2}$ \quad
\textbf{Purity Kamene Kavwele}$^{1}$ \quad
\textbf{Marian Klose}$^{3}$ \\
\vspace{0.2em}
\textbf{Piyush Kumar}$^{1}$ \quad
\textbf{Rams\'es J.~S\'anchez}$^{4}$ \quad
\textbf{Darius A.~Faroughy}$^{5}$ \\
\vspace{0.4em}
{\normalsize
$^{1}$University of Potsdam, Germany \quad
$^{2}$Technical University Berlin, Germany \quad
$^{3}$Freie Universit\"at Berlin, Germany \\
$^{4}$University of Bonn, Germany \quad
$^{5}$NHETC, Rutgers University
}
}

\begin{document}
\maketitle

\begin{abstract}
 We introduce Prior-Fitted Functional Flows, a generative foundation model for pharmacokinetics that enables zero-shot population synthesis and individual forecasting without manual parameter tuning. We learn functional vector fields, explicitly conditioned on the sparse, irregular data of an entire study population. This enables the generation of coherent virtual cohorts as well as forecasting of partially observed patient trajectories with calibrated uncertainty. We construct a new open-access literature corpus to inform our priors, and demonstrate state-of-the-art predictive accuracy on extensive real-world datasets.
 %Validated against extensive real-world datasets, our method demonstrates state-of-the-art predictive accuracy.
\end{abstract}

\section{Introduction}
\label{sec:intro}
%- What is important
%- what is done
%- what is wrong
%- how do we do solve it

Pharmacokinetics (PK) studies the dynamics of how the human body processes a drug.
In a typical early-development setting, each subject receives a single dose, and the goal is to characterize how systemic drug concentration unfolds over time.
This characterization requires not only representing temporal dynamics, but also capturing \textit{heterogeneity} because, even under identical doses, different individuals can exhibit markedly different profiles, due to factors such as physiology, disease state, and co-medications.

The key difficulty is that these dynamics must be inferred from sparse and noisy measurements collected across a \textit{study}: a cohort of individuals, each observed at an irregular, and often subject-specific set of time points. 
Operationally, one would like to (i) predict the concentration profile of a new individual, or (ii) forecast the unobserved future of a patient from only a few early measurements.
PK formalizes this setting by treating each measured concentration as a noisy sample of an underlying, patient-specific, and continuous-time \textit{concentration function} generated by a dynamical system~\citep{gibaldi1982pharmacokinetics}.
In the population setting, this is most often expressed through nonlinear mixed-effects (NLME) models, which decompose variability into fixed effects shared across the cohort, and random effects capturing individual deviations~\citep{lavielle2014}.
From this viewpoint, the fundamental inferential object is a \textit{distribution over plausible concentration functions}, conditioned on study information and (optionally) partial observations.

Recent work has begun to approximate such concentration functions with deep generative models, such as neural ODEs~\citep{lu2021deep}. 
However, PK datasets are not only sparse and noisy, but also small (especially in Phase I and many Phase II settings). 
Training a high-capacity model from scratch for each new dataset is therefore both statistically fragile and operationally expensive. 
To bridge the gap between expressive modeling and data scarcity, we turn to \textit{amortized inference}.

Amortization shifts the computational cost of inference from repeated, dataset-specific optimization to a single, up-front \textit{pretraining} phase across large and diverse datasets --- typically synthetically generated from \textit{mechanistic priors}. 
The heterogeneous pretraining distributions encourage models to learn \textit{reusable inference algorithms} that map a context (\textit{e.g.}, the study and partial observations) directly to a target (posterior) distribution.
This paradigm underlies prior-fitted networks (PFNs) for tabular data~\citep{muller2022transformers, hollmann2023tabpfn, hollmann2025accurate}, foundation inference models (FIMs) for system identification~\citep{berghaus2024foundation, fim_sde, berghaus2026incontext, mauel2026foundation}, and many foundation models for time-series imputation~\citep{seifner2025zeroshot} and forecasting~\citep{dooley2024forecastpfn, bhethanabhotla2024mamba4cast, hemmer2025true}.

\texttt{AICMET}~\citep{marin2025amortized} represents a first step in this direction for PK.
It was pretrained on synthetic studies generated from stochastic priors on compartmental models, and performs \textit{zero-shot posterior estimation} over concentration functions using a hierarchical mixed-effects representation.
The latter is implemented through neural-process-style latent variables~\citep{garnelo2018neural}.
A limitation of this design is the reliance on a compressed representation of the study context, which can lead to underfitting when the context is complex~\citep{kim2019attentive}, and to uncertainty that is not sharply calibrated~\citep{wang2025bridge}.

In this work, we dispense from explicit mixed-effect parametrizations and instead learn the distribution over concentration functions directly in function space.
We introduce \textit{Prior-fitted Functional Flows} (PFFs), which leverages conditional optimal-transport (OT) flow matching~\citep{kerrigan2024dynamic} to transport a simple reference measure over functions (a Gaussian process) into the conditional distribution over drug-concentration functions. 
The construction is conditional at two levels: first at the level of the study (\textit{i.e.}, the cohort-level context), and second, at the level of a patient’s early measurements for forecasting.
We therefore make the following contributions:
\begin{enumerate}        
    \item We release an open-access reference dataset automatically mined from clinical bioequivalence studies, comprising inferred pharmacokinetic parameters and synthetic concentration–time samples generated from them. We use this resource to calibrate the physiological plausibility of our synthetic priors.
    %We extend the \texttt{AICMET} simulator to induce a broader prior over physiologically plausible PK profiles, and validate this prior against a large collection of bioequivalence studies.    
    \item We introduce \texttt{PFF}, a functional flow-matching model pre-trained on simulated PK studies to amortize inference of concentration profiles from sparse data. It parameterizes the transport vector field with continuum-attention neural operators~\citep{calvello2024continuum}, enabling queries at arbitrary time points, and achieves accurate zero-shot predictions, outperforming NLME and \texttt{AICMET}.

    \item We formalize study-level conditioning and individual-level partial-observation conditioning within a conditional optimal-transport framework: study context modulates the vector field via attention, while individual forecasting transports probability mass only in the unobserved future using triangular maps~\citep{kerrigan2024dynamic, chemseddine2025conditional}.
    %\item We introduce \texttt{PFF}, the first functional flow-matching model \textit{pretrained} on simulated PK studies to \textit{amortize} the inference of concentration profiles from sparse study data. The model parametrizes the transport vector field using \textit{continuum-attention neural operators}~\citep{calvello2024continuum}, which can be queried at arbitrary time points.
    %
    %We show that \texttt{PFF} generates accurate concentration profiles in \textit{zero-shot} mode (that is, in a single forward pass) given the context, and outperforms both classical NLME baselines and \texttt{AICMET}.
    
    %\item We distinguish study-level conditioning from patient-level partial-observation conditioning, and instantiate both within a conditional optimal-transport framework. Study information enters the learnable vector field through attention mechanism, while patient-level forecasting follows the conditional OT viewpoint: the observed portion of a patient’s trajectory is held fixed, and probability mass is transported only in the unobserved future component via \textit{triangular maps}~\citep{kerrigan2024dynamic}.
\end{enumerate}

\section{Related Work}
\label{sec:related-work}

\textbf{Classical PK and NLME Modelling}. Population PK is typically formulated with nonlinear mixed-effects (NLME) models, where a compartmental ODE is shared across subjects and parameters decompose into fixed (population) and random (individual) effects~\citep{lavielle2014}. Recent ML extensions include GP and neural mixed-effects formulations, including mixed-effects neural ODEs \cite{kipnis2025metabeta}, but these approaches are often developed for data-rich settings rather than the small, sparse studies common in early drug development. And different from us, focus on learning parameters of mechanistic models \cite{haggstrom2025simulation}.

\textbf{Amortized Inference for PK}. Simulation-based amortization reduces the cost of inference by pretraining neural estimators on simulated data~\citep{pmlr-v235-arruda24a}. While such pipelines can accelerate individual-level inference, study-level fitting is typically still repeated per dataset. \texttt{AICMET} amortizes further by mapping study context directly to posterior-predictive concentration functions in a zero-shot manner~\citep{marin2025amortized}, but retains limitations of neural-process-style latent-variable modeling.

\textbf{Generative Modelling for Continuous Functions}. Function-space generative models learn continuous-time transport maps between distributions over trajectories, including diffusion-based and flow-matching approaches~\citep{bilovs2023modeling,kerrigan2023functional,kollovieh2024flow,jahn2025trajectory}. Our method builds most directly on the dynamic conditional optimal-transport perspective for functional flow matching~\citep{kerrigan2024dynamic}.

\section{Preliminaries}
\label{sec:preliminaries}
\begin{figure}[h]
    \centering
    \includegraphics[width=\linewidth]{figures/NotationIllustration_FlowMatching.pdf}
    \caption{The model conditions on a population study (green) and dosing schedules (yellow). Based on partial observations (blue) of a new patient, it generates plausible continuations (black) to produce future forecasts (red).
        }
    \label{fig:graphical-model}
\end{figure}

In this section, we introduce the notation and formalize the amortized conditional generation problem we aim to solve. 
We then review the optimal-transport (OT) and flow-matching tools we build on, culminating in the conditional OT formulation that underlies our construction.

\subsection{Problem Definition}

We formalize pharmacokinetic (PK) data as sparse, noisy observations of latent, continuous-time concentration functions, and cast inference as learning a conditional probability measure over such functions.

\textbf{Study data}.
A pharmacokinetic study consists of noisy concentration measurements collected from a cohort of $I$ individuals $\mathcal{S}=\{\mathcal{D}^1,\dots,\mathcal{D}^I\}$.
For individual $i$, we write $\mathcal{D}^{i} = (\textbf{u}^i, \boldsymbol{\tau}^{i}, \mathbf{y}^{i})$, where 
$\mathbf{u}^i=(a^i,r^i)$ is the dosing descriptor, with dose amount $a^i\in\mathbb{R}^+$ and administration route $r^i\in\{1,\dots,R\}$.
The vector $\boldsymbol{\tau}^{i} = (\tau^{i}_{1}, \dots, \tau^{i}_{T_i})$ contains the (irregular, subject-specific) observation times, and
$\mathbf{y}^{i} = (y^{i}_{1}, \dots, y^{i}_{T_i})$ contains the measured (\textit{e.g.}, plasma) concentrations,
with $T_i$ the number of available measurements. 
% Thus, each $\mathcal{D}^{i}$ defines a \textit{concentration trajectory}, and $\mathcal{S}$ a set of trajectories. 
%
Note that the observation times need not be shared across individuals, reflecting the sparse and irregular sampling typical in early-development studies.

\textbf{Latent concentration functions}.
Let $\mathcal{T} = [0, T]$ denote the observation window of interest, with horizon time $T$. 
% {\color{red}[ $\mathcal{T}$ is a subintervall? or a set of points? in either case we would need $\subset$ instead of $\in$]}.
%
We model each individual's concentration trajectory as a \textit{latent function} $f: \mathcal{T} \rightarrow \mathbb{R}$, and interpret the measurements $\mathbf{y}^i$ as noisy, irregular samples of $f$ at times $\boldsymbol{\tau}^i$. 
To work in a well-defined function space, let $\mathcal{F}$ be a real, separable Hilbert space of functions 
equipped with its Borel $\sigma$-algebra $\mathcal{B}(\mathcal{F})$ and norm $||\cdot\|_\mathcal{F}$.
Let us also denote with $\PM(\F)$ the set of Borel probability measures on $\mathcal{F}$.

\textbf{Conditional probability measures}.
PK inference is inherently conditional. Predictions for a new individual depend on the study $\mathcal{S}$, dosing information, and possibly a small number of early measurements for that individual. 
For now, we collect all such information into a single context variable $c \in \mathscr{C}$, where $\mathscr{C}$ is an abstract context space. 
The inferential object of interest is therefore a \textit{conditional probability measure} over concentration functions $\nu(\cdot|c) \in \PM(\F)$. 
This measure should capture both the uncertainty due to sparse and noisy observations, and the heterogeneity across individuals implied by the context.

\textbf{The amortized conditional generation problem}.
Let $\eta \in \PM(\F)$ be a tractable reference measure over functions (\textit{e.g.}, a Gaussian process).
Our goal is to learn a single conditional generator that, given $\eta$ and a context $c$, produces samples distributed according to $\nu(\cdot|c)$.
Concretely, we seek a measurable map $T^\star: \mathscr{C} \times \mathcal{F}\rightarrow \mathcal{F}$ such that, for (almost) every context $c \in \mathscr{C}$ we have
%\begin{equation}
$(T^\star(c, \cdot))_{\#} \eta = \nu(\cdot | c).$ %\label{eq:conditional-generation-problem}%\end{equation}
In Section~\ref{sec:conditional-ot}, we introduce conditional optimal transport on product spaces, and the class of triangular maps that formalize the conditional generation problem. %(Eq.~\ref{eq:conditional-generation-problem}).
Then, in Section~\ref{sec:model}, we instantiate $c$ to reflect the PK settings of interest --- that is, study-conditioned generation and, for forecasting, conditioning on partial observations of a new patient. 
\begin{figure*}[t]
    \centering
    \includegraphics[width=0.8\linewidth]{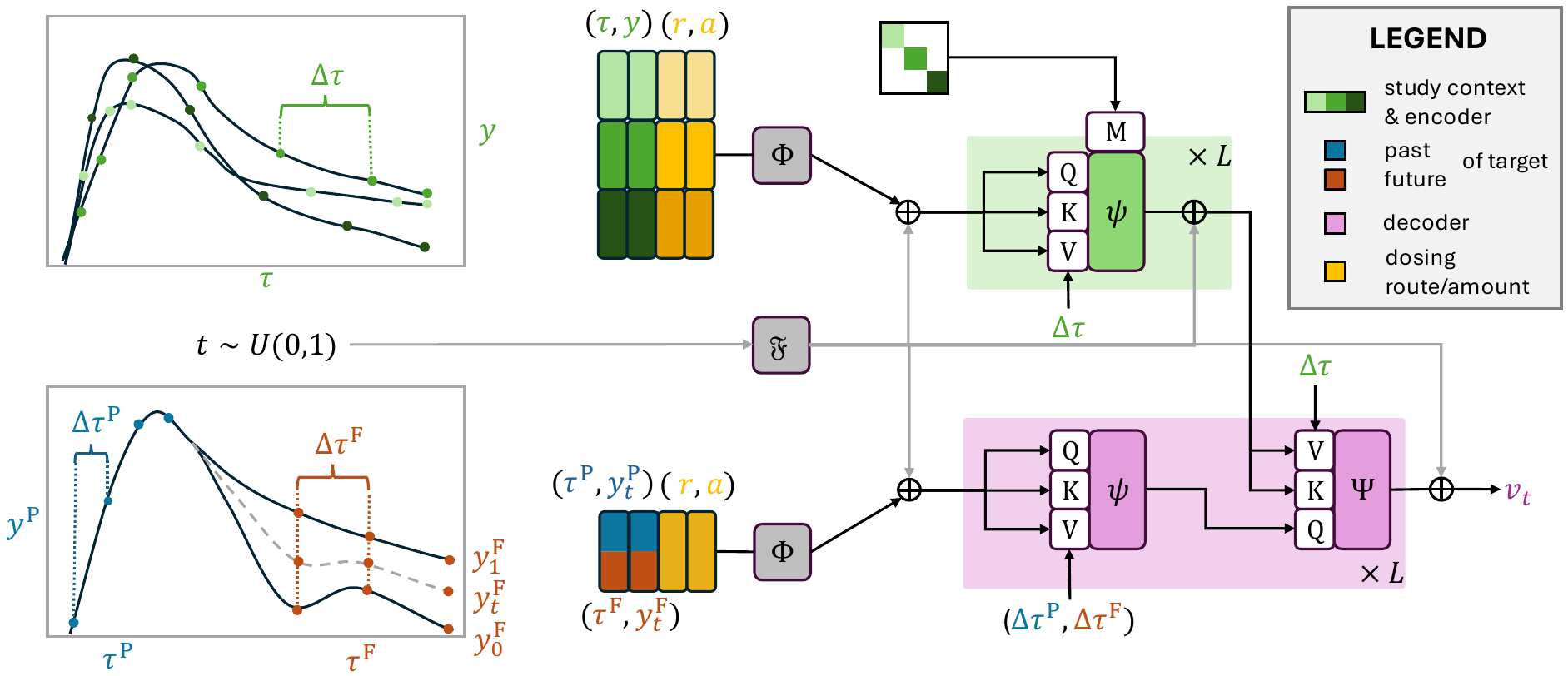}
    \vspace{-0.4em}
    \caption{
    Top left: study context consists of time series of different individuals. Bottom left: data for a target individual are split into past observations $(\tau^{\tt P},y^{\tt P})$ and future observations $(\tau^{\tt F},y_1^{\tt F})$. With a uniformly distributed interpolation parameter $t$, $y_t^{\tt F}$ is obtained by interpolating between observed data $y_1^{\tt F}$ and a sample $y_0^{\tt F}$ from a Gaussian  conditional path. Right: model architecture consisting of a study context encoder (top) and decoder (bottom). Abbreviations: $\tau$: study time; $y$ observed plasma concentration; $r$: route of administration; $a$: dosing amount; $\Phi$, multilayer perceptron; $\mathfrak{F}$: Fourier features; $\psi$/$\Psi$: self-/cross-attention; Q/K/V/M/L: query/key/value/block diagonal mask/layers.
%    First we encode $\mathcal{S}$ with a self attention that respects the permutational invariance of the representation through masking. We interpolate between $\*x_{\tt F}$ and  $y_{\tt F}$ to obtain $y^{\tt F}_t$. representations at both the individual and study levels. Our transformer-based decoder embeds the encoder representations alongside dose information. Finally, we define functional queries that allow us to evaluate the predictive distribution at any target time $\tau$. By introducing $\mathbf{z}_i$ and $\mathbf{z}_s$, we can model fixed and random effects, enabling a population-aware, individualized characterization of dynamics.
}
    \label{fig:architecture}
\end{figure*}

\subsection{OT and Flow Matching}

Our model builds on a standard link between optimal transport (OT) and flow-based generative modeling. 
%
%OT provides a principled way to move probability mass from a simple \textit{source} distribution to a complex \textit{target} distribution, while flow models realize this movement via an ODE driven by a learned velocity field~\citep{lipman2022flow}. 
%
%The Benamou–Brenier formulation makes this connection explicit by characterizing OT as the minimum kinetic energy needed to transport one distribution into another over continuous time.

\textbf{Unconditional OT}.
Let $\eta,\nu\in\PM(\F)$ be probability measures on the (separable Hilbert) function space $\F$. 
In the Monge formulation, OT seeks a measurable transport map $T:\F\to\F$ that pushes $\eta$ to $\nu$ (\textit{i.e}, $T_{\#}\eta=\nu$) while minimizing squared displacements. 
This yields the squared 2-Wasserstein distance
\begin{equation}
W_2^2(\eta,\nu) = \inf_{T:\F\to\F} \int_{\F} \|f - T(f)\|_{\F}^2\, d\eta(f),
\label{eq:w2-functional}
\end{equation}
where the infimum is taken over all $T$ which fulfill $T_{\#}\eta=\nu$.

The same quantity admits the dynamical Benamou-Brenier  formulation~\citep{benamou2000computational}:
\begin{equation}
W_2^2(\eta,\nu) = \min_{(\gamma_t,v_t)} \int_0^1\int_{\F} \|v_t(f)\|_{\F}^2\, d\gamma_t(f)\, dt,
\label{eq:benamou_brenier}
\end{equation}
subject to the boundary conditions $\gamma_0=\eta$, $\gamma_1=\nu$, and the continuity equation
\begin{equation}
\partial_t\gamma_t+\mathrm{div}(v_t\gamma_t)=0.
\label{eq:continuity}
\end{equation}
Here $(\gamma_t)_{t\in[0,1]}\subset\PM(\F)$ is a time-indexed path of measures, and $v_t:\F\to\F$ is a time-dependent \textit{vector field}.
Intuitively, $v_t$ specifies how ``particles'' $f$ move through $\F$, so that their resulting path measure $(\gamma_t)_{t\in[0,1]}$ evolves from $\eta$ to $\nu$. Thus the transport map $T$ is parameterized through the vector field $v_t$.
%
% This is precisely the viewpoint adopted by flow-based generative models. Draw $f(0) \sim \eta$ and evolve $f(t)$ according to $\dot f (t) = v_t(f(t))$ to obtain $f(1) \sim \nu$.

 The above two formulations are also known as static and dynamic optimal transport, respectively. In practice, directly solving the Benamou–Brenier minimization problem (Eq. \ref{eq:benamou_brenier}) is intractable, as it requires optimizing over the space of all possible measure paths and vector fields. Flow Matching circumvents this by using the fact that any vector field satisfying the continuity equation (Eq. \ref{eq:continuity}) generates a valid transport (though not ad hoc one of minimal costs).  More specifically, Flow matching replaces direct OT optimization with a supervised regression problem, in which one (i) defines a family of simple conditional interpolation paths between paired samples $f_0 \sim \eta$ and $f_1 \sim \nu$, 
(ii) derives the corresponding conditional velocity field explicitly, and 
(iii) learns a parametric model $v^\theta_t(\cdot)$ to match this vector field.
While in (i) any coupling $\rho$ works, choosing an OT-aligned coupling (e.g., via mini-batch OT) is crucial; it ensures the interpolation paths are straight, which minimizes the kinetic energy and aligns the Flow Matching objective with the Benamou–Brenier objective ~\citep{pooladian2023multisample, tong2023improving}. We explain the functional version of Flow Matching in more detail in the following section.
%
%We also note conditional extensions developed in subsequent work~\citep{chemseddine2025conditional}.

\textbf{Functional Flow Matching.}~\citep{kerrigan2023functional}. 
Let $f_0\sim\eta$ and $f_1\sim\nu$, and let $z=(f_0,f_1)\sim\rho$ for some coupling $\rho\in\Pi(\eta,\nu)$ (later to be chosen as minibatch-OT). 
To address step (i) from the previous section, Functional Flow Matching defines a Gaussian conditional path $\gamma_t(\cdot| z)$, which interpolates between source and target samples: \begin{equation}
\gamma_t(\,\cdot \mid z) = \mathcal{N}\!\bigl((1-t) f_0 + t f_1,\; \Sigma\bigr), 
\label{eq:fm-cond-path}
\end{equation}
where $\Sigma$ is a trace-class covariance operator on $\F$ \citep{da2014stochastic}.
This conditional path has the advantage that it allows us to derive the associated conditional vector field $v_t(\cdot| z)$ corresponding to step (ii) explicitly:
\begin{equation}
v_t(\cdot \mid z) = f_1 - f_0.
\label{eq:fm-cond-vf}
\end{equation}
While these conditional objects describe the trajectory of individual sample pairs, the marginal path $\gamma_t$ describes the evolution of the full probability density from $\eta$ to $\nu$. It is obtained by mixing the conditional path over the coupling,
\begin{equation}
\gamma_t =\int \gamma_t(\,\cdot \mid z)\, d\rho(z),
\label{eq:fm-marg-path}
\end{equation}
and the corresponding marginal vector field can be written as
\begin{equation}
v_t(\cdot)
=
\int v_t(\cdot \mid z)\,
\frac{d\gamma_t(\,\cdot \mid z)}{d\gamma_t}\,
d\rho(z).
\label{eq:fm-marg-vf}
\end{equation}
Crucially, the goal is to learn this marginal field $v_t$, which drives the global transport.
To realize step (iii), one learns a parametric vector field $v^\theta_t(f)$ by solving the regression problem for the conditional vector field (Eq.~\ref{eq:fm-cond-vf})
\begin{equation*}
\mathcal{L}(\theta)=\mathbb{E}_{t\sim\mathcal{U}[0,1]}\,
\mathbb{E}_{z\sim\rho}\,
\mathbb{E}_{f\sim\gamma_t(\cdot\mid z)}
\bigl[\|v^\theta_t(f)- (f_1-f_0)\|_{\F}^2\bigr].
\label{eq:fm-loss}
\end{equation*}
Theoretical results guarantee that minimizing this objective has the same gradients as the (intractable) mean-squared error to the marginal field
 $v_t$ (Eq.~\ref{eq:fm-marg-vf}) \citep{tong2023improving}. 

\subsection{Conditional Optimal Transport in Function Space}
\label{sec:conditional-ot}
The previous section described Flow Matching in the \textit{unconditional} setting. However, our problem is inherently conditional: we observe a context $c\in \CM$ (e.g., patient covariates, dosing regimen) and seek to sample from a target conditional measure $\nu(\cdot\mid c)\in\PM(\F)$. 
Rather than solving a separate generative problem for every possible context, we formalize this as Conditional Optimal Transport. We seek a single \textit{triangular} transport map on the product space $\CM\times\F$ that generalizes across contexts, effectively amortizing the inference. As a warning, it is important to distinguish the notion of conditioning used in Flow Matching (conditioning on the path endpoints $z=(f_0, f_1)$ to define the vector field) from the conditioning on an external context $c \in \mathscr{C}$ (e.g., patient covariates).

%\subsection{Conditional Optimal Transport in Function Space}
%\label{sec:conditional-ot}

%The previous section described Flow Matching in the \textit{unconditional} setting.
%
%Now, we turn to the conditional problem, in which we observe a context $c \in \mathscr{C}$ and seek to sample from a (target) conditional measure $\nu(\cdot\mid c) \in \PM(\F)$.
%
%Conditional optimal transport formalizes this by learning a single transport map on the product space $\mathscr{C} \times \F$ that generalizes across contexts.

\paragraph{Conditional OT.}
Let $\mu \in \PM(\mathscr{C})$ denote a measure over contexts. We consider joint measures on $\mathscr{C} \times \F$ whose marginal on $\mathscr{C}$ equals $\mu$,
\begin{equation}
\PM_{\mu}(\mathscr{C} \times \F)
=
\Bigl\{\pi \in \PM(\mathscr{C} \times \F) : \pi_{\mathscr{C}} = \mu \Bigr\},
\label{eq:Pmu_def}
\end{equation}
where $\pi_{\mathscr{C}}$ denotes the $\mathscr{C}$-marginal of $\pi$. We take as reference joint measure the product $\eta
=
\mu \otimes \eta_{\F},$ where $\eta_{\F} \in \PM(\F)$ is an easy-to-sample base measure over functions. A target joint measure $\nu \in \PM_{\mu}(\mathscr{C} \times \F)$ is characterized by its conditional distributions $\nu(\cdot\mid c)$.

For conditional OT, we seek \textit{triangular} transport maps $T: \mathscr{C} \times \F \to \mathscr{C} \times \F$.
These maps leave the context component unchanged, formally defined as:
\begin{equation}
T(c,f) = \bigl(c,\, T_{\F}(c,f)\bigr),
\label{eq:triangular_map}
\end{equation}
where $T_{\F}: \mathscr{C} \times \F \to \F$.
If $T$ is triangular and satisfies the global pushforward constraint $T_{\#} \eta = \nu$, then the \textit{fiberwise pushforward} identity holds for $\mu$-almost every context $c \in \mathscr{C}$:
\begin{equation}
\bigl(T_{\F}(c,\cdot)\bigr)_{\#} \eta_{\F}
=
\nu(\cdot \mid c).
\label{eq:fiberwise_pushforward}
\end{equation}
This formulation effectively amortizes the transport problem: learning a single global map $T$ solves the conditional generation problem for all contexts simultaneously.

\paragraph{Triangular Functional Flows.}
In the dynamic Benamou--Brenier framework (Eq.~\ref{eq:benamou_brenier}), a triangular transport map corresponds to a flow generated by a \textit{triangular vector field} $v_t: \mathscr{C} \times \mathcal{F} \to \mathscr{C} \times \mathcal{F}$.
The context $c$ is static , i.e. it represents fixed attributes that do not evolve during the generative process, which implies that its time derivative must be zero. This imposes a block structure (triangularity) on the vector field:
\begin{equation}
v_t(c,f) = \begin{pmatrix} 0 \\ v_t^\mathcal{F}(c,f) \end{pmatrix},
\label{eq:triangular_vf}
\end{equation}
where $v_t^\mathcal{F}: \mathscr{C} \times \mathcal{F} \to \mathcal{F}$ drives the evolution of the function $f$ conditioned on $c$.
Consequently, the learning problem reduces to estimating the functional component $v_t^\mathcal{F}$. We achieve this by parameterizing $v_t^\mathcal{F}$ with a neural network and minimizing the Flow Matching objective (Eq.~\ref{eq:fm-loss}), treating $c$ as an additional conditioning input to the network.

\section{Prior-Fitted Functional Flows}
\label{sec:model}
\label{ssec:latent-generative}

% In this section, we introduce our latent-variable model. In simple terms, our model Prior Fitted Functional Flow, treats the Mix Effects models by conditioning on the study dataset $\mathcal{S}$ and extends the time series model by introducing the measurement variable $p$. Now, instead of proposing a generative model over the time series $\*Y^n$, we do a generative model over $\*Y^n_f$ conditioned on $\*Y^n_p$ where if $p= 0$ we recover $\emptyset$. 

In this section, we introduce \textit{Prior-Fitted Functional Flows} (\texttt{PFF}), a methodology for \textit{zero-shot inference} of distributions (measures) over concentration functions from sparse PK studies. 
The core idea is to learn a context-conditioned transport map in function space that pushes a simple reference measure (a Gaussian process) to the \textit{posterior predictive distribution} implied by a study context. 
%
%Transport is represented through a time-dependent, conditional vector field on function space, so that sampling reduces to drawing a function from the reference measure, and integrating an ODE driven by the learned field.
This is achieved by training a neural operator to approximate the triangular vector field derived in Section~\ref{sec:conditional-ot}.
Rather than fitting scarce real-world datasets, we generate synthetic studies from a broad prior over physiologically plausible pharmacokinetic models.
Crucially, while the data relies on a simulator, the training of the vector field is based on Flow Matching and therefore is \textit{simulation-free}.
%

%What is more, training is entirely simulation-based. 
%
%We generate synthetic studies from a broad prior over physiologically plausible compartmental PK models, and train a neural operator to approximate the conditional vector field. 
%
This yields an \textit{amortized} model that can be applied off-the-shelf: given a new study (and, when available, a few early measurements from a new individual), \texttt{PFF} produces samples from a predictive distribution over full concentration trajectories in a single forward pass (plus ODE solve).

% A central design choice is to distinguish two levels of conditioning. First, study-level conditioning encodes population information—how profiles vary across individuals under the study’s dosing and sampling patterns—and enters the vector field through attention-based conditioning on the study set. Second, partial-observation conditioning enforces consistency with early measurements for a new individual and is implemented through the conditional optimal-transport viewpoint via triangular (block) maps, which preserve the conditioning component while transporting uncertainty in the remaining degrees of freedom. We postpone the exact instantiation of this structure to the subsequent subsections; for now, it suffices to view both mechanisms as different ways of constructing the context variable c from the preliminaries.

This approach is close in spirit to prior-fitted networks in that, rather than fitting a new model per dataset, we train once on a simulator-defined prior and learn an inference procedure that outputs the posterior predictive distribution over functions directly from context. 
We thus refer to our model as Prior-Fitted Functional Flows.
\subsection{Pretraining Prior Distribution}
\label{sec:pretraining-prior}
We construct a large-scale synthetic pretraining corpus using a hierarchical generative model $P^{sim}$ that combines standard compartmental ODE systems with stochastic, time-varying parameters. This setup allows for the simulation of diverse study designs, including variable dosing routes, compartment structures, and irregular sampling schedules based on physiological timescales. A core contribution of this work is the rigorous validation of this generative prior against a large-scale dataset of pharmacokinetic parameters automatically mined from clinical literature, ensuring that our synthetic training data faithfully reflects real-world physiological variability. The full dataset and generation pipeline are made publicly available, with comprehensive details provided in Appendix~\ref{app:literature-extraction}.

%We simulate synthetic PK studies by sampling study-specific designs (number of individuals, compartments, dose, and route), evolving latent compartment amounts with standard absorption–distribution–elimination Compartemental Models ODE, and endowing each individual’s log-kinetic parameters (and volume) with stochastic variability via independent Ornstein-Uhlenbeck processes whose hyperparameters are drawn from empirically calibrated priors. Observations are generated by selecting irregular, physiologically informed sampling times based on characteristic scales (peak time and half-life) and sub-sampling to match empirical study designs. The resulting hierarchical construction defines an explicit joint generative model over study datasets, serving as the simulator used for pretraining and as the reference prior matched by our latent-variable formulation (details deferred to the appendix).

\begin{figure*}[h]
    \centering
    % Single row (3 panels)
    \begin{minipage}[b]{0.28\textwidth}
        \centering
        (a)
        \adjustbox{trim=0pt 0pt 0pt 0pt, clip}{
            \includegraphics[width=\linewidth]{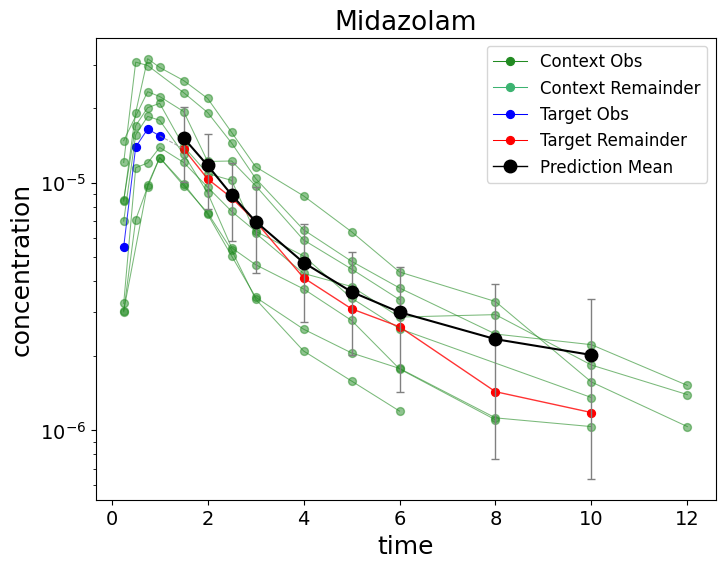}
        }
    \end{minipage}
    \hspace*{0.015\textwidth}
    \begin{minipage}[b]{0.28\textwidth}
        \centering
        (b)
        \adjustbox{trim=0pt 0pt 0pt 0pt, clip}{
            \includegraphics[width=\linewidth]{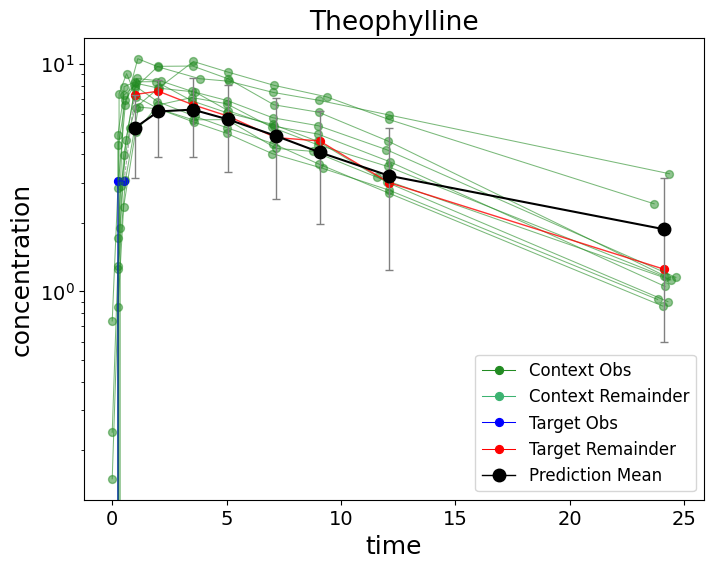}
        }
    \end{minipage}
    \hspace*{0.015\textwidth}
    \begin{minipage}[b]{0.28\textwidth}
        \centering
        (c)
        \adjustbox{trim=0pt 0pt 0pt 0pt, clip}{
            \includegraphics[width=\linewidth]{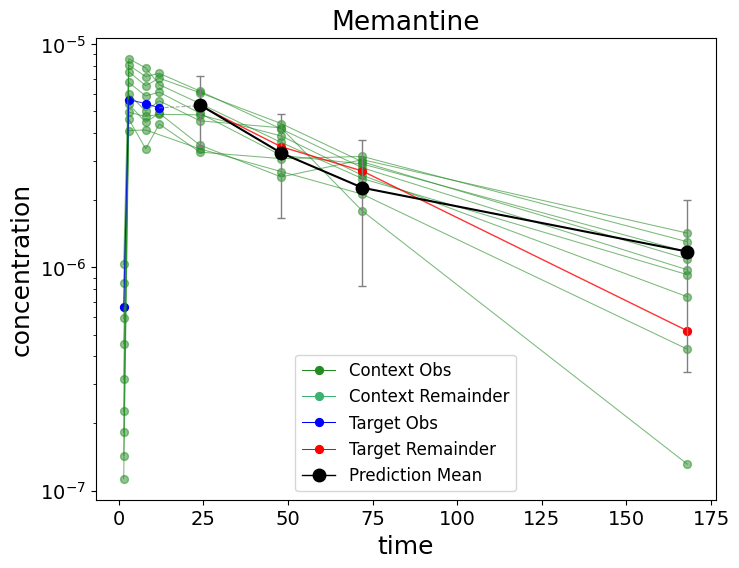}
        }
    \end{minipage}

    \caption{Predictive plots for three representative compounds: (a) Midazolam, (b) Theophylline, (c) Memantine.  Conditioned on the population study context (green), the model uses the partial observations (blue) of a new patient to sample the forecasts (red) from the predictive distribution, with the prediction mean (black) shown for reference.}
    \label{fig:prediction_row}
\end{figure*}

\subsection{Conditional Functional Flow Matching Model}
\label{sec:model-cffm}

We now instantiate the abstract conditional generation problem from Section~\ref{sec:preliminaries} in the PK setting, and describe the conditional flow-matching construction used by Prior-Fitted Functional Flows (\texttt{PFF}).
The key objects are: (i) a \emph{simulator-induced} target measure over concentration trajectories, and (ii) a single learnable vector field in function space whose induced flow maps a tractable reference measure to the target in zero-shot mode.
Crucially, \texttt{PFF} is trained once and used for both population synthesis and forecasting. 
The two modes differ only in the context provided at inference time, and in the corresponding choice of reference measure.

Thus, our methodology distinguishes two conditioning levels
\begin{enumerate}
\item \textit{Study-level conditioning} on $\mathcal S$, which is always present and encodes population variability and sampling patterns. In \texttt{PFF}, $\mathcal S$ enters the vector field via attention-based mechanisms.
\item \textit{Individual-level conditioning} on an optional prefix of measurements for a new individual.
When a prefix is available, \texttt{PFF} performs forecasting by transporting uncertainty only in the unobserved continuation. When no prefix is available, \texttt{PFF} reduces to population synthesis.
\end{enumerate}

Accordingly, we define a context space $\CM = \mathscr{S} \times \mathscr{O}$, 
where $\mathscr{S}$ denotes the set of studies, and $\mathscr{O}$ collects the information available for a new individual. 
In \textit{population synthesis} we take $\mathscr{O}=\{\varnothing\}$ (no individual measurements). 
In \textit{forecasting} we use a nontrivial element $o\in\mathscr{O}$ defined below.

\textbf{A new individual and prefix split}.
We distinguish the observed study $\mathcal S$ from a new \textit{virtual} individual $\mathcal D^{\rm n} = (\mathbf u^{\rm n},\boldsymbol{\tau}^{\rm n},\mathbf y^{\rm n})$,
for which we may observe a short prefix of measurements. 
We introduce a split index $p \in \{0,\dots,P\}$,
which partitions the observations into a revealed \textit{past} and an unobserved \textit{future}. Writing $\mathbf{y}^{\rm n}=(\mathbf y^{\rm n}_{\tt P},\mathbf y^{\rm n}_{\tt F})$ and $\boldsymbol{\tau}^{\rm n}=(\boldsymbol{\tau}^{\rm n}_{\tt P},\boldsymbol{\tau}^{\rm n}_{\tt F})$,
we define
\[
\mathbf y^{\rm n}_{\tt P}=(y_1^{\rm {\rm n}},\dots,y_p^{\rm n}),\quad
\mathbf y^{\rm n}_{\tt F}=(y_{p+1}^{\rm {\rm n}},\dots,y_{T_{\rm n}}^{\rm n}),
\]
and similarly for $\boldsymbol{\tau}^{\rm n}_{\tt P},\boldsymbol{\tau}^{\rm n}_{\tt F}$.  Setting $p=0$ corresponds to \text{population synthesis}, so $o=\varnothing$ and $c=(\mathcal S,\varnothing)$. For $p>0$, we perform \textit{forecasting}, with $o = (\mathbf u^{\rm n},\boldsymbol{\tau}^{\rm n}_{\tt P},\mathbf y^{\rm n}_{\tt P},p) \in \mathscr{O}$, and $c \;=\; (\mathcal S,o)\in\CM$. In what follows, we drop the superscript $n$ for new individuals whenever the meaning is clear from context.

\textbf{Past-future decomposition in function space}.
Let $\mathcal T=[0,T]$ and let $\F= C([0,T])$ be a Hilbert space of continuous concentration trajectories.
This choice reflects the physical fact that PK concentration trajectories are continuous in time, and it allows us to speak meaningfully about values at a split time.
For a given prefix length $p$, we define the split time $\tau_p$ and the corresponding intervals $\mathcal T_{\tt P}=[0,\tau_p]$, and $\mathcal T_{\tt F}=[\tau_p,T]$.
We represent the full trajectory as a pair
\[
f \equiv (f_{\tt P},f_{\tt F}),
\qquad
f_{\tt P}:\mathcal T_{\tt P}\to\mathbb R,
\qquad
f_{\tt F}:\mathcal T_{\F}\to\mathbb R,
\]
and we identify the full function space with the product $\F = \F_{\tt P}\times\F_{\tt F},$
where $\F_{\tt P}$ and $\F_{\tt F}$ are Hilbert spaces of functions on $\mathcal T_{\tt P}$ and $\mathcal T_{\tt F}$, respectively. 
To ensure the pair represents a single continuous trajectory, we impose compatibility at intersection, i.e., $f_{\tt P}(\tau_p) = f_{\tt F}(\tau_p)$.
Intuitively, $f_{\tt P}$ is the revealed prefix segment and $f_{\tt F}$ is the unobserved continuation.

What is more, we define $\tau_0 = 0$ so that, when $p=0$, the ``past'' interval collapses, and the ``future'' interval corresponds to the full domain. 
This makes population synthesis a special case of the same construction.

\textbf{Simulator-induced target measure}.

Our simulator $P^{\mathrm{sim}}$ first samples latent kinetic parameters (physics) and study designs (dosing, covariates, sampling times).
Based on these, it generates continuous concentration curves $f$ and compiles them into discrete studies $\mathcal{S}$ (which contain discrete observations of multiple curves).
Crucially, we cannot query the simulator to generate curves for a specific, pre-defined study $\mathcal{S}$; we are restricted to the studies it stochastically generates.
Our learning task is therefore to use generated joint pairs $(\mathcal{S}, f)$ to approximate the conditional measure $\nu^{\mathrm{sim}}(\,\cdot\mid \mathcal S)$, enabling us to generalize to arbitrary studies later.
To extend this framework to forecasting, where an individual is partially observed, we decompose the trajectory as $f=(f_{\tt P},f_{\tt F})$. This joint law defines the \textit{past marginal}
$\nu^{\mathrm{sim}}_{\tt P}(\,\cdot\mid\mathcal S)\ \in\ \PM(\F_{\tt P})$,
and a corresponding \textit{future conditional}  $\nu^{\mathrm{sim}}_{\tt F}(\,\cdot\mid\mathcal{S},f_{\tt P})\ \in\ \PM(\F_{\tt F}).$ Informally, $\nu^{\mathrm{sim}}_{\tt F}$ describes plausible continuations compatible with a given prefix $f_{\tt P}$.
In practice, accessing $f_{\tt P}$ directly is impossible, and the individual-level context enters through the observations $o = (\boldsymbol{\tau}^n_{\tt P},\mathbf y^n_{\tt P}, \mathbf{u}^n)$.
The ojective is to learn to sample from these target conditionals, i.e. $\nu^{\mathrm{sim}}(\cdot \mid \mathcal{S})$ for synthesis and $\nu^{\mathrm{sim}}_{\tt F}(\cdot \mid \mathcal{S}, o)$ for forecasting, for \textit{any} observed context.

%
%Our simulator $P^{\mathrm{sim}}$ defines a broad prior over physiologically plausible compartmental PK models and induces, for each study $\mathcal S$, a measure over full concentration functions for a new individual $\nu^{\mathrm{sim}}(\,\cdot\mid \mathcal S)\ \in\ \PM(\F)$.

%By sampling $f\sim \nu^{\mathrm{sim}}(\cdot\mid\mathcal S)$ and writing $f=(f_{\tt P},f_{\tt F})$, we obtain the induced joint measure
%
%$\nu^{\mathrm{sim}}_{\tt P, \tt F}(\,\cdot\mid\mathcal S)\ \in\ \PM(\F_{\tt P}\times\F_{\tt F})$.
%
%This joint law defines the \textit{past marginal}
%$\nu^{\mathrm{sim}}_{\tt P}(\,\cdot\mid\mathcal S)\ \in\ \PM(\F_{\tt P})$,
%
%and a corresponding \textit{future conditional} 
%\begin{equation}
%\nu^{\mathrm{sim}}_{\tt F}(\,\cdot\mid\mathcal{S},f_{\tt P})\ \in\ \PM(\F_{\tt F}),
%\end{equation}
%so that, informally, $\nu^{\mathrm{sim}}_{\tt F}$ describes plausible continuations compatible with a given prefix function $f_{\tt P}$ under the PK prior.
%
%In practice, we do \textit{not} provide $f_{\tt P}$ to the model directly when forecasting. 
%
%The individual-level context enters only through the discrete prefix observations $(\boldsymbol{\tau}^n_{\tt P},\mathbf y^n_{\tt P}, \mathbf{u}^n)$ contained in $o$.

\textbf{Reference measure (unified synthesis and forecasting)}.
To train a single model that covers both modes, we use a reference measure on $\F_{\tt F}$ that depends on whether a prefix is available.
Let $\mathrm{GP}_{\mathcal T_{\tt F}}(\cdot)$ denote a Gaussian-process prior on $\mathcal T_{\tt F}$, and let $\mathrm{GP}_{\mathcal T_{\tt F}}(\cdot\mid \mathbf u,\mathbf y_{\tt P})$ denote the corresponding GP posterior restricted to $\mathcal T_{\tt F}$.
We define
\begin{equation}
\eta_{\tt F}(\cdot\mid o, p)
:=
\begin{cases}
\mathrm{GP}_{\mathcal T_{\tt F}}\bigl(\,\cdot\mid \mathbf u,\mathbf y_{\tt P}\bigr)
&\text{if} \, \, \, \, p>0,\\[4pt]
\mathrm{GP}_{\mathcal T_{\tt F}}(\,\cdot)
&\text{if} \, \, \, \, p=0,
\end{cases}
\label{eq:reference-measure}
\end{equation}
such that $\eta_{\tt F}(\cdot\mid o, p)\in\PM(\F_{\tt F})$.
Thus, in population synthesis ($p=0$) the flow starts from a GP prior over full trajectories (since $\F_{\tt F}\equiv\F$), whereas in forecasting ($p>0$) it starts from a GP posterior that extrapolates from the observed prefix.
In both cases, we use the same trainable vector field conditioned on the study $\mathcal S$. 
The only additional change is whether the input includes an individual prefix, which alters the reference measure and the conditioning context.

\begin{algorithm}[t]
\caption{Training of Prior-Fitted Functional Flows}
\label{alg:pff-training}
\DontPrintSemicolon
\SetKwInOut{KwIn}{Input} \SetKwInOut{KwOut}{Return}

\KwIn{Simulator $P^{\mathrm{sim}}$, Network $v_\theta$, Step size $\eta$.}
\KwOut{Trained parameters $\theta$}

\While{not converged}{
    $(\mathcal{S}, \mathbf{y}^{\rm n}, \boldsymbol{\tau}^{\rm n}) \sim P^{\mathrm{sim}}$ \tcp*{Sample from Simulator}
    $p \sim \mathcal{U}\{0, \dots, |\boldsymbol{\tau}^{\rm n}|-1\}$ \tcp*{ Sample Split} $\mathbf{y}^{\rm n} \to (\mathbf{y}_{\tt P}, \mathbf{y}_{\tt F})$ and $\boldsymbol{\tau}^{\rm n} \to (\boldsymbol{\tau}_{\tt P}, \boldsymbol{\tau}_{\tt F})$ \tcp*{Partition Data}
   $c = (\mathscr{S},\mathbf{u}^{\rm n}, \boldsymbol{\tau}_{\tt P}, \mathbf{y}_{\tt P})$ \tcp*{Set Context}
    $\mathbf{x}_{\tt F} \sim \text{GP}(\boldsymbol{\tau}_{\tt F} \mid \mathbf{y}_{\tt P}, \boldsymbol{\tau}_{\tt P})$    \tcp*{Sample Reference}
    $\mathbf{z}_0 = (\mathbf{y}_{\tt P}, \mathbf{x}_{\tt F})$, 
    $\mathbf{z}_1 = (\mathbf{y}_{\tt P}, \mathbf{y}_{\tt F})$ \tcp*{Set Start and End}

    $t \sim \mathcal{U}(0,1)$, $\mathbf{z}_t \sim \mathcal{N}(t \mathbf{z}_1 + (1-t)\mathbf{z}_0, \sigma_{\min}^2 I)$\;
    
    $u_t = (\mathbf{0}, \mathbf{y}_{\tt F} - \mathbf{x}_{\tt F})$ \tcp*{Target Velocity}
    
    $\mathcal{L}(\theta) = \| \mathbf{M}_{\tt F} \odot (v_\theta(\mathbf{z}_t, t, \mathcal{S}) - u_t) \|^2$ \tcp*{Loss}
    
    $\theta \leftarrow \theta - \eta \nabla_\theta \mathcal{L}(\theta)$\;
}
\end{algorithm}

%\begin{algorithm}[t]
%\caption{Training of Prior Fitted Functional Flow Matching}
%\label{alg:tsflow-uncond}
%\DontPrintSemicolon
%\SetKwInOut{KwIn}{Input}
%\SetKwInOut{KwOut}{Return}

%\KwIn{meta prior $P^{sim}$, network $v^\theta$, step size $\eta$.}
%\KwOut{$v^\theta$}

%\While{not converged}{
%    $ (\mathbf y^{\rm n}_{\tt F}, \boldsymbol{\tau}^{\rm n}_{\tt F},\boldsymbol{\tau}^{\rm n}_{\tt P},\mathbf y^{\rm n}_{\tt P}, \mathbf{u}^{\rm n},\mathcal{S})  = (\mathbf y^{\rm n}_{\tt F}, \boldsymbol{\tau}^{\rm n}_{\tt F}, o,\mathcal{S}) \sim P^{sim}$\;
 
%    $ f_{\tt F} \sim \eta_{\tt F}(\cdot\mid o, p) \text{ \quad } \*x_{\tt F} = f_{\tt F}(\boldsymbol{\tau}^{\rm n}_{\tt F})  $\;
%    $(x_0,y) \leftarrow \mathrm{OT}(x_0,x_1)$\;
%    $t \sim \mathcal{U}(0,1)$\;
%    $\mathbf{z}_t \sim \mathcal{N}\!\bigl(t \, \*z_1+(1-t)\*z_0,\;\sigma_{\min}^2 I\bigr) $\;
%    $\mathcal{L}(\theta) \leftarrow \left\|
%\*M_p \odot v^\theta_t(\mathbf{z}_t, \mathcal{S})
%-
% (\mathbf{y}^n_{\tt F}-\mathbf{x}_{\tt F})
%\right\|^2$\;
%    $\theta \leftarrow \theta - \eta\,\nabla_\theta \mathcal{L}(\theta)$\;
%}
%\end{algorithm}

\textbf{Conditional flow matching for PK}.
The data generation process of Section~\ref{sec:pretraining-prior} yields samples $\mathcal{D}^{\rm n}, \mathcal{S} \sim P^{\mathrm{sim}}$. 
Let us now sample $p \sim \text{Uniform}(0, \dots, P)$ and introduce the \textit{finite-dimensional} flow variables 
$\mathbf{z}_0 = (\mathbf{y}_{\tt P}^{\rm n}, \mathbf{x}_{\tt F})$ and $\mathbf{z}_1 = (\mathbf{y}_{\tt P}^{\rm n}, \mathbf{y}^{\rm n}_{\tt F})$. 
Here $\mathbf{x}_{\tt F} = \left(f_{\tt F}(\tau_{p+1}^{\rm n}), \dots, f_{\tt F}(\tau_{T_{\rm n}}^{\rm n}) \right)$, where $f_{\tt F} \sim \eta_{\tt F}$ is sampled from the reference measure (Eq.~\ref{eq:reference-measure}).

Similar to Eq.~\ref{eq:fm-cond-path}, we now introduce the (finite-dimensional) Gaussian conditional path 
\begin{equation}
\*z_t \sim p_t(\*z \mid \*z_0, \*z_1) =\mathcal{N}\!\bigl(t \, \*z_1+(1-t)\*z_0,\;\sigma_{\min}^2 I\bigr),
\end{equation}
and the associated target, triangular vector field
$
v_t(\*z \mid \*x_{\tt F},\*y_{\tt F}^{\rm n}) = (\mathbf{0}, \mathbf{y}_{\tt F}^{\rm n}- \mathbf{x}_{\tt F}).
$
Using this construction, we train a neural operator vector field
$v^\theta_t(\mathbf{z}, c)$ by minimizing the conditional flow matching loss
\begin{equation*}
\mathcal{L}(\theta)
=
\mathbb{E}_{t,p,\mathbf{x}_{\tt F},\mathbf{y}^{\rm n}_{\tt F}}
\!\left[
\left\|
\*M_p \odot v^\theta_t(\mathbf{z}_t, c)
-
 (\mathbf{y}^{\rm n}_{\tt F}-\mathbf{x}_{\tt F})
\right\|^2
\right],
\end{equation*}
where $\*M_p$ is a prefix-dependent mask that zeros out the ``past'' of the learnable vector field.

\begin{table*}[h!]
    \caption{\textbf{Comparison of Log-RMSE Across Models.} For each compound, log-RMSE is reported for baseline models (NLME and NODE-PK), and the newly proposed AICMET model including its ablations. The best-performing model for each compound is highlighted in \textbf{bold}. }\vskip 5mm
    \label{tab:main_results}
    
    \centering
    \renewcommand{\arraystretch}{1.2} % Adjust row height
    \rowcolors{3}{orange!15}{white} % Alternating row colors
    \setlength{\tabcolsep}{5pt} % Reduce column spacing
    \resizebox{0.7\linewidth}{!}{ % Adjust width to fit column
\begin{tabular}{llllll}
\toprule
 & NLME & Transformer & NODE & AICMET & FlowPK \\
drug &  &  &  &  &  \\
\midrule
caffeine & $0.294 \pm 0.294$ & $0.451 \pm 0.258$ & $0.432 \pm 0.209$ & $0.399 \pm 0.205$ & $\mathbf{0.174 \pm 0.086}$ \\
dextromethorphan & $0.752 \pm 0.752$ & $\mathbf{0.395 \pm 0.163}$ & $0.605 \pm 0.248$ & $0.595 \pm 0.364$ & $0.963 \pm 0.426$ \\
digoxin & $\mathbf{0.300 \pm 0.300}$ & $0.683 \pm 0.242$ & $0.661 \pm 0.315$ & $0.691 \pm 0.464$ & $0.375 \pm 0.216$ \\
indometacin & $0.582 \pm 0.582$ & $0.429 \pm 0.328$ & $0.350 \pm 0.291$ & $0.536 \pm 0.203$ & $\mathbf{0.245 \pm 0.170}$ \\
memantine & $\mathbf{0.343 \pm 0.343}$ & $0.513 \pm 0.335$ & $0.554 \pm 0.386$ & $0.475 \pm 0.341$ & $0.404 \pm 0.388$ \\
midazolam & $0.596 \pm 0.596$ & $0.288 \pm 0.173$ & $0.352 \pm 0.224$ & $\mathbf{0.264 \pm 0.151}$ & $0.295 \pm 0.300$ \\
omeprazole & $1.300 \pm 1.300$ & $0.842 \pm 0.400$ & $1.003 \pm 0.618$ & $0.936 \pm 0.942$ & $\mathbf{0.593 \pm 0.161}$ \\
paracetamol & $\mathbf{0.299 \pm 0.299}$ & $0.447 \pm 0.274$ & $0.582 \pm 0.302$ & $0.352 \pm 0.187$ & $0.309 \pm 0.220$ \\
repaglinide & $0.479 \pm 0.479$ & $0.709 \pm 0.268$ & $0.693 \pm 0.290$ & $\mathbf{0.339 \pm 0.311}$ & $0.681 \pm 0.155$ \\
rosuvastatin & $\mathbf{0.383 \pm 0.383}$ & $0.502 \pm 0.205$ & $0.535 \pm 0.303$ & $0.558 \pm 0.286$ & $0.495 \pm 0.336$ \\
theophylline & $0.732 \pm 0.732$ & $0.356 \pm 0.141$ & $0.327 \pm 0.144$ & $0.274 \pm 0.109$ & $\mathbf{0.184 \pm 0.110}$ \\
tolbutamide & $0.723 \pm 0.723$ & $0.678 \pm 0.339$ & $0.578 \pm 0.263$ & $0.517 \pm 0.254$ & $\mathbf{0.421 \pm 0.214}$ \\
1-hydroxy-midazolam &  & $0.567 \pm 0.327$ & $0.737 \pm 0.381$ & $0.491 \pm 0.309$ & $\mathbf{0.422 \pm 0.205}$ \\
4-hydroxy-tolbutamide &  & $0.354 \pm 0.174$ & $0.353 \pm 0.182$ & $0.335 \pm 0.164$ & $\mathbf{0.293 \pm 0.153}$ \\
5-hydroxy-omeprazole &  & $1.286 \pm 0.472$ & $1.474 \pm 0.745$ & $1.188 \pm 0.451$ & $\mathbf{1.026 \pm 0.356}$ \\
dextrorphan &  & $0.733 \pm 0.445$ & $0.834 \pm 0.284$ & $\mathbf{0.563 \pm 0.287}$ & $0.585 \pm 0.487$ \\
hydroxy-repaglinide &  & $0.069 \pm 0.217$ & $\mathbf{0.049 \pm 0.156}$ & $0.076 \pm 0.240$ & $0.070 \pm 0.221$ \\
omeprazole sulfone &  & $\mathbf{1.054 \pm 0.297}$ & $1.390 \pm 0.488$ & $1.082 \pm 0.509$ & $1.156 \pm 0.355$ \\
paracetamol glucuronide &  & $0.357 \pm 0.192$ & $0.297 \pm 0.169$ & $0.307 \pm 0.184$ & $\mathbf{0.265 \pm 0.114}$ \\
paraxanthine &  & $0.413 \pm 0.181$ & $0.287 \pm 0.135$ & $0.312 \pm 0.138$ & $\mathbf{0.273 \pm 0.143}$ \\
\bottomrule
\end{tabular}
    }
\end{table*}

\textbf{Vector-field parameterization}.
We parameterize the transport through a time-dependent triangular velocity field $v^{\theta}:\ [0,1]\times \mathscr{S} \times \F_{{\tt F}} \to \F_{{\tt F}}$. The model is an \emph{operator encoder--decoder transformer} that mirrors the standard sequence-to-sequence architecture of \citet{vaswani2017attention}. A \emph{context encoder} maps the context $\mathscr{C}$ to a set of subject-level representations using multi-head self-attention, with block-diagonal masking to prevent information flow across subjects. A \emph{target decoder} takes the flow-interpolated target individual $\mathbf{z}_t$ and applies non-causal self-attention over its past and future representations, followed by cross-attention blocks over the study-encoder outputs. The observation times are treated as positional encoding. One key departure from the standard transformer architecture is that both encoder and decoder operate on discretized functions over observation times that are subject-specific and not aligned to a common regular time grid. All attention operations are therefore replaced by their \emph{continuum operator} counterparts \citep{calvello2024continuum}, which correct for the grid irregularities via a trapezoidal quadrature approximation. For further details on the architecture see appendix~\ref{app:architecture}.

\begin{figure}[t]
    \centering
    \begin{minipage}[b]{0.4\textwidth}
        \centering
        (a)
        \adjustbox{trim=0pt 0pt 0pt 0pt, clip}{
            \includegraphics[width=\linewidth]{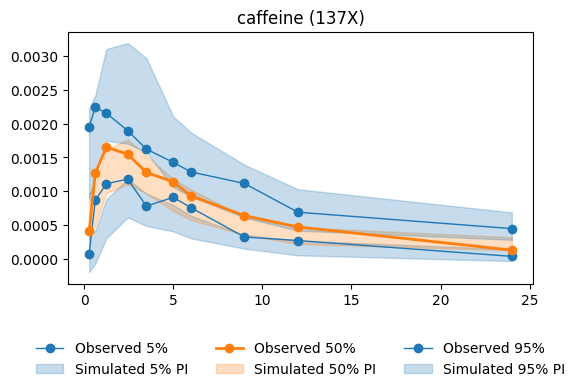}
        }
    \end{minipage}
    \hspace*{0.02\textwidth}
    \begin{minipage}[b]{0.4\textwidth}
        \centering
        (b)
        \adjustbox{trim=0pt 0pt 0pt 0pt, clip}{
            \includegraphics[width=\linewidth]{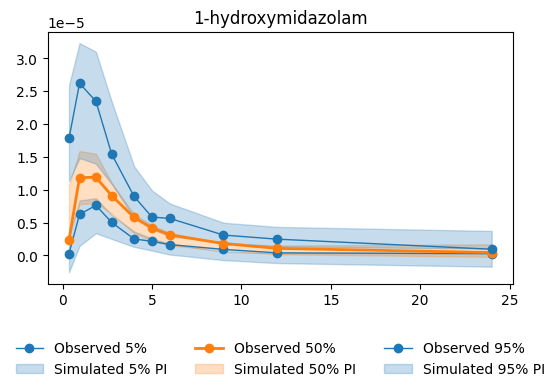}
        }
    \end{minipage}
    \caption{Visual predictive checks (VPCs) for (a) caffeine and (b) 1-hydroxy-midazolam. Each panel compares observed concentrations to simulation-derived prediction intervals.}
    \label{fig:vpc_only}
\end{figure}

\section{Results}
\label{sec:results}

In this section, we present the evaluation dataset and baselines and afterwards discuss our main results.
\subsection{Datasets}
\label{sec:data}
Evaluation data were obtained from the open-access PK-DB repository, which aggregates pharmacokinetic measurements from clinical and pre-clinical studies~\citep{grzegorzewski2021pkdb}.
From PK-DB we curated plasma concentration--time measurements from a Phase~I study in healthy volunteers investigating the safety and dosing of a combination of drugs and their metabolites~\citep{lenuzza2016safety}.
In total, the dataset comprises 18 compounds (parent drugs and primary metabolites) with 2019 valid concentration--time observations.
Sampling frequencies vary substantially both within and across compounds, resulting in irregularly sampled time series.
Each compound is measured in 10 individuals, with up to 15 observations per individual. 
We also included two additional datasets on indometacin and theophylline pharmacokinetics (\texttt{Indometh} and \texttt{Theoph} from the R \texttt{datasets} package) \citep{Rsoftware}.

\subsection{Baselines and Ablations}
\label{sec:baselines}

We compare our proposed Prior-Fitted Functional Flows (PFF) against a classical pharmacometric baseline, nonlinear mixed-effects modeling (NLME), as well as two neural baselines: NODE-PK and the AICMET Transformer. The NLME baseline is fitted study-by-study for each compound using standard estimation procedures (one- or two-compartment models with first-order absorption, selected by AIC), including inter-individual variability and combined error models. Metabolites were excluded as they lack explicit dosing inputs. In contrast, all neural models are pre-trained on our synthetic datasets and evaluated in-context on empirical studies without fine-tuning. NODE-PK extends latent neural ODEs~\citep{rubanova2019latent} with explicit dosing handling via an RNN encoder. But note that it is not capable of study-level aggregation~\citep{lu2021deep}. The AICMET Transformer~\citep{marin2025amortized} serves as a hierarchical neural process baseline, capturing both population-level and individual-level effects via separate encoding variables. 
\subsection{Metrics}

\paragraph{Forecasting evaluation} The performance of the past-conditioned models were assessed using a leave-one-individual-out setting: predictions for subject $i$ were conditioned on the study context $\mathcal{S}$, dosing $\mathbf{u}^i$, and the first four PK samples $\mathcal{D}^i_{\text{C}}$, then evaluated on the remaining samples $\mathcal{D}^i_{\text{T}}$ using log-RMSE.\\

\paragraph{Generative evaluation} We evaluate time series sample quality in two separate settings: an \emph{empirical} setting and a \emph{synthetic} setting. 
In both cases, the goal is to compare a reference set of trajectories against a corresponding set of trajectories generated by the model under the same study-level conditioning, but the way these two sets are constructed differs across the two settings.
For the empirical evaluation, we apply a leave-one-out protocol across all empirical studies and all available drugs. 
In each study, one individual is removed from the study context, the model is conditioned on the remaining individuals, and one trajectory is sampled for the held-out individual. 
Repeating this procedure over all individuals in all studies yields a generated empirical set whose cardinality matches that of the full empirical dataset, since each empirical individual is held out and resampled exactly once. 
The resulting model-generated trajectories are then pooled across studies and compared against the pooled empirical trajectories from the original dataset.
For the synthetic evaluation, no leave-one-out construction is needed. 
Instead, each synthetic study provides a study context together with a reference population of target individuals sampled directly from the simulator. 
Conditioned on the same context, the model generates a matching synthetic population, and these model-generated trajectories are compared directly against the corresponding oracle trajectories.
After constructing these empirical and synthetic comparison pairs, we quantify discrepancy in both settings using two complementary two-sample metrics: 
(i) the area under the ROC curve (AUC) of a classifier trained to distinguish reference from generated trajectories, where values closer to $0.5$ indicate better sample fidelity, and 
(ii) a maximum mean discrepancy (MMD) computed with the signature kernel of \citet{kiraly2019kernels}, which measures distributional discrepancy between the two sets of irregular trajectories in path space.

Furthermore, the generative capability of the models were assessed via visual predictive checks (VPC), comparing simulation percentiles against observed concentrations~\citep{holford2005}. Full implementation details are provided in the Supplementary Material.

\begin{table}[t]
\centering
\caption{\textbf{Sample Fidelity Scores} Lower is better for MMD, while AUC values closer to $0.5$ indicate better agreement.}
\label{tab:sample_fidelity}
\begin{tabular}{lcc}
\toprule
\textbf{Model} & \textbf{AUC} & \textbf{MMD} \\
\midrule
\multicolumn{3}{c}{\textbf{Empirical}} \\
\cmidrule(lr){1-3}
AICMET  & 	
0.577 & 	
-0.001 \\
FlowPK & 	
\textbf{0.567} & 	
\textbf{-0.003} \\
\addlinespace
\multicolumn{3}{c}{\textbf{Synthetic}} \\
\cmidrule(lr){1-3}
AICMET  & 0.541 & 	
0.002 \\
FlowPK & \textbf{0.508} & 	
\textbf{0.001} \\
\bottomrule
\end{tabular}
\end{table}

\subsection{Discussion of numerical results}

Table~\ref{tab:main_results} summarizes the quantitative forecasting performance. \texttt{PFF} achieves the lowest or on par log-RMSE across the evaluated compounds in the majority of cases. This predictive accuracy and robust uncertainty calibration are illustrated in Figure~\ref{fig:prediction_row}, which illustrates inferred future trajectories for three partially observed individuals. Furthermore, Figure~\ref{fig:vpc_only} confirms the model's generative capacity for zero-shot population synthesis; the Visual Predictive Checks demonstrate that \texttt{PFF}'s simulated percentiles tightly align with the empirical data distributions of entire study cohorts.
In Table~\ref{tab:sample_fidelity} we present the results for generated sample fidelity compared against the other generative model for Pharmacokinetic AICMET. Our model outperforms in both metrics the previous competing generative model.

\section{Conclusion}
\label{sec:conclusion}

We introduced Prior-Fitted Functional Flows (PFF), a generative foundation model for pharmacokinetics that enables zero-shot population synthesis and individual forecasting without manual parameter tuning. By learning functional vector fields conditioned on entire study populations, PFF captures complex, multi-modal distributions that classical methods often miss. Crucial to our methodology is the construction of a large-scale, open-access literature benchmark, which we used to rigorously validate the physiological realism of our synthetic priors.
\paragraph{Limitations.} While PFF demonstrates state-of-the-art accuracy, it currently relies on purely synthetic pre-training (though informed by our systematic literature-mined corpus). Incorporating real-world clinical data into the pre-training would further enhance its ability to capture the pathological variability of specific disease populations. Furthermore, our current validation focuses on single-dose studies; extending the framework to handle complex, irregular multi-dosing remains a key direction for future work. Finally, as with all generative models, there is a risk of generating plausible but incorrect trajectories in extremely sparse data regimes, necessitating careful uncertainty quantification in clinical applications.
\subsection*{Acknowledgements}
\label{sec:scknowledgement}

This research has been funded by the Deutsche Forschungsgemeinschaft (DFG) --- Project-ID 318763901 --- SFB1294; and by the Federal Ministry of Education and Research of Germany and the state of North-Rhine Westphalia as part of the Lamarr Institute for Machine Learning and Artificial Intelligence. DAF is supported by DOE grant
DOE-SC0010008. This research used resources of the National Energy Research Scientific Computing Center, a DOE Office of
Science User Facility supported by the Office of Science of the U.S. Department of Energy under Contract No. DE-AC02-05CH11231 using NERSC award HEPERCAP0027491

% References
\bibliography{uai2026-template}

\newpage

\onecolumn

\title{Prior Fitted Functional Flows \\(Supplementary Material)}
\maketitle

\section{Generative Data Model and Empirical Validation}

In this section, we introduce a hierarchical data generation framework that combines mechanistic compartmental pharmacokinetic (PK) systems with stochastic, time-varying parameters. We utilize this model to generate a large-scale synthetic training dataset. We build upon the probabilistic generative framework introduced in the AICMET model \cite{marin2025amortized}, described in Section \ref{sec:pk-systems}. While the original framework relied on internal meta-analyses for calibration, a core contribution of this work is the rigorous validation of the generator against a large-scale dataset of pharmacokinetic parameters automatically mined from clinical literature (Section \ref{app:literature-extraction}).

\subsection{Mechanistic Generative Process}
\label{sec:pk-systems}
The data generation process is hierarchical: we first sample a study configuration (population priors), then sample individual subjects, and finally generate their continuous drug concentration trajectories .
\paragraph{Compartmental Pharmacokinetic Systems.}
Each simulated study consists of $I$ individuals, where each individual is characterized by a time-varying latent state $\mathbf{X}(t) \in \mathbb{R}^{2+P}$ that represents the drug amount in the gut, central (c), and $P$ peripheral (per) compartments. That is
\begin{equation}
    \mathbf{X}(t)= \Bigl(X_{\text{gut}},\,X_{c},\,X_{\text{per},1}, \dots, X_{\text{per},P} \Bigr)^{\!\top}.
\end{equation}
%
% \[
% \mathbf{X}(t)=
%   \Bigl(
%     X_{\text{gut}},\,
%     X_{c},\,
%     X_{\text{per},1},\dots,X_{\text{per},P}
%   \Bigr)^{\!\top}\in\mathbb{R}^{2+P}
% \]
%
which solves the ODE system
\begin{align}
\dot X_{\text{gut}}      &= -k_{a}\,X_{\text{gut}}, \nonumber\\
\dot X_{c}               &= k_{a}X_{\text{gut}}
                           -\Bigl(k_{e}+\sum_{j=1}^{P}k^{+}_{j}\Bigr)X_{c}
                           +\sum_{j=1}^{P}k^{-}_{j}X_{\text{per},j}, \nonumber\\
\dot X_{\text{per},j}    &= k^{+}_{j}X_{c}-k^{-}_{j}X_{\text{per},j},                
\label{eq:pk-odes}                           
\end{align}
with $j$ running from $1$ to $P$.
The initial condition is determined by the dose and the dosing route (i.e.,~$X_{u_2=c}(0)=u_1$ for intravenous dosing and $X_{u_2=\text{gut}}(0) = u_1$ for oral dosing), while all other states are initialized at zero.
The number of individuals, the number of peripheral compartments, as well as the dose and dosing route, are randomly drawn for each simulated study.

\paragraph{Stochastic parameter model.}
To capture realistic inter- and intra-individual variability beyond standard parametric families, we model the log-kinetic parameters $\theta_i(t)=\log ( V^i, k^i_a,  k^i_e, , k^{i+}_1,  k^{i-}_1,..., k^{i+}_P, k^{i-}_P)$ as independent Ornstein-Uhlenbeck (OU) processes :
\begin{equation}
d\theta_{i,k}(t)=
  -\lambda_{k}\!\left(\theta_{i,k}(t)-\mu_{i,k}\right)\,dt
  +\sigma_{k}\,dW_{i,k}(t),
\label{eq:ou}
\end{equation}
initialized at $\theta_{i,k}(0)\sim \mathcal{N}(\mu_{i,k},\sigma_k^2/(2\lambda_k))$ to ensure stationarity.
% , i.e., $\theta_{i,k}(t)\sim \mathcal{N}(\mu_{i,k},\sigma^2/(2\lambda))$ for $t\ge 0$.

The hyperparameters governing these processes, namely the long-term means $\mu_{i,k}$, reversion speeds $\lambda_k$, and volatilities $\sigma_k$, are drawn hierarchically from broad uniform priors calibrated to cover physiologically plausible ranges, see Section \ref{app:literature-extraction}. Finally, to mimic clinical trial conditions, sparse observations are generated by sampling the continuous trajectories at irregular intervals derived from characteristic timescales $T_{\tt peak}$ and $T_{\tt half}$ and applying proportional measurement noise, see \cite{marin2025amortized}.

\subsection{Automated Extraction of Pharmacokinetic Parameters from the Literature}
\label{app:literature-extraction}

To validate that the generative model produces physiologically realistic concentration-time profiles, we constructed a reference dataset of real-world pharmacokinetic (PK) parameters.
Since raw subject-level data is rarely available, we utilized a Large Language Model (LLM) pipeline to mine reported summary statistics from \textit{bioequivalence (BE) studies} published in the open-access biomedical literature. 
BE studies typically compare a \emph{test} formulation (e.g., a generic) against a \emph{reference} formulation (e.g., an originator) under defined dosing conditions to enable abbreviated regulatory approval if predefined PK criteria relative to the reference product are met. 
These studies are a particularly well-suited reference because they follow standardized protocols, typically enroll healthy volunteers under controlled conditions, and are required by regulatory agencies to report a consistent set of PK summary statistics.
Furthermore, dense sampling ensures that individual PK profiles are well characterized, allowing for the derivation of reliable PK metrics~\citep{FDA2021Bioequivalence}.

\subsubsection{Data Acquisition Pipeline}

\textbf{Step 1: Literature Retrieval}

We first constructed a candidate corpus ($N=200$) of BE-related publications from PubMed Central (PMC)~\citep{pmc_database} by querying the web interface using the search term \texttt{"bioequivalence"[title]}.  
The resulting records contained bibliographic metadata, including PubMed identifiers (PMIDs) and PMC identifiers (PMCIDs).  
For each of the top 200 retrieved articles, we used the \texttt{PubMedFetcher} function from \texttt{metapub} 0.6.4~\citep{most_metapub} to obtain the structured abstract via the PMID and the \texttt{Entrez.efetch} function from \texttt{Biopython} 1.86~\citep{biopython} to retrieve the complete full-text XML from PMC via the PMCID.
Full-text XML was parsed with \texttt{BeautifulSoup} 4.13.4~\citep{beautifulsoup4_2025} to obtain plain text.

\textbf{Step 2: LLM-Based Triage}
\label{app:triage}

Not all retrieved papers reported primary PK trial data; many were reviews, commentaries, meta-analyses, or methodological discussions about BE aspects.
To exclude these from the database in a time- and cost-efficient manner, we employed an LLM agent (\texttt{gpt-5-nano}) via OpenAI's \texttt{Responses API} in Python 3.12.6 receiving the title and abstract of the paper as user message.
The agent was instructed via a detailed system message to classify papers as \textit{Suitable} (True) only if they reported a single, distinct bioequivalence trial.
Machine-readable outputs were enforced using a \texttt{Pydantic} 2.10.5~\citep{pydantic_validation} class in combination with the \texttt{structured outputs} feature of the \texttt{Responses API}, which guarantees that the model returns a  Boolean flag rather than free text.
This step discarded 86 papers and reduced the corpus to a set of 114 clinical BE trials suitable for further extraction.

\textbf{Step 3: Structured Data Extraction}
\label{app:extraction}

For each article that passed triage, we prompted the complete full-body text to the LLM with a different system prompt, instructing it to extract the following information:
\begin{itemize}
    \item \textbf{Study-level metadata:} Compound name, number of subjects enrolled, and year of publication.
    \item \textbf{Dosing Protocol:} Route of administration (e.g., oral), formulation type (e.g., tablet, capsule), and amount and unit of administered dose.
    \item \textbf{Sampling design:} Time of the last blood sample collected ($t_{\text{last}}$).
    \item \textbf{PK characteristics:} Central tendency and dispersion of the \textit{test} formulation for:
    \begin{itemize}
        \item $C_{\max}$ (Maximum concentration)
        \item $T_{\max}$ (Time to maximum concentration)
        \item $\text{AUC}_{0-t}$ (Area under the concentration--time curve up to the last measured time point)
%        \item $\text{AUC}_{0-\infty}$ (Area under the concentration--time curve extrapolated to infinity)
    \end{itemize}
\end{itemize}

\begin{figure}[htp!]
    \centering
    \includegraphics[width=\textwidth]{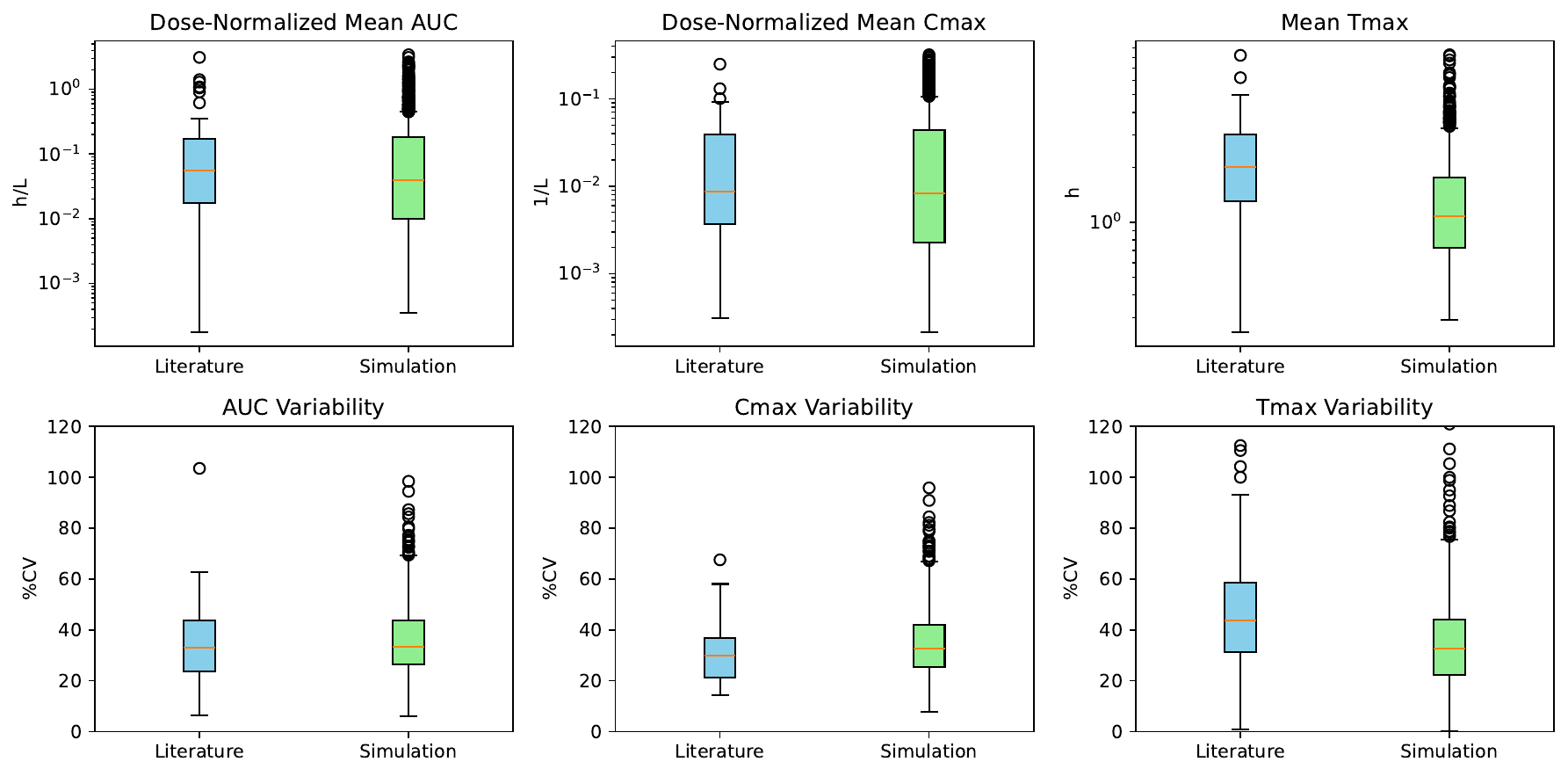}
    \caption{Comparison of three key PK characteristics (Mean/\%CV) between literature and 1000 simulated datasets.}
    \label{fig:benchmark}
\end{figure}

To ensure machine-readable consistency, we again utilized \texttt{structured outputs} enforced via nested \texttt{Pydantic} models, yielding one JSON record for each paper.
Where both fasting and fed-state results were reported, the LLM was instructed to only take fasting values into account.
The total API costs for this LLM-based workflow across all 200 screened and 114 classified papers was \$0.31 USD.

\subsection{Post-Processing}
\label{app:postprocessing}

The extracted JSON records were post-processed into a final analysis literature dataset via the following steps:

\begin{itemize}
    \item Amongst the 114 BE trials, we kept only those 85 corresponding to oral administration\footnote{While our framework is capable of simulating intravenous administration as well, a focus on oral dosing allowed for a more meaningful comparison of Cmax and Tmax};
    \item Units were harmonized across studies and records with incomplete or implausible values (e.g., $T_{\max} > t_\text{last}$ or missing units) were excluded;
    \item $\mathrm{AUC}_{0-t}$ and $C_{\max}$ were dose-normalized;
    \item The dispersion metrics were converted to coefficients of variation (\%CV), which was either directly used if reported or calculated from central tendency and absolute dispersion metrics (using mean/median interchangeably and assuming range=4~SD). 
\end{itemize}

The resulting dataset provided empirical reference distributions for dose-normalized $\mathrm{AUC}_{0-t}$, dose-normalized $C_{\max}$ and $T_{\max}$ across a diverse set of compounds and doses, which we used to benchmark the marginal distributions produced by our stochastic generative model.

\subsubsection{Benchmark of data generative model against literature}
\label{app:benchmark}

The parameters of the stochastic data-generating model were matched to the PK metrics obtained from the literature, as shown in Fig.~\ref{fig:benchmark}.

\begin{table}[h]
\centering
\caption{Meta-study configuration parameters used for synthetic PK meta-study generation.}
\label{tab:meta_study_config}
\begin{tabular}{lc}
\toprule
\textbf{Parameter} & \textbf{Value / Range} \\
\midrule

\multicolumn{2}{l}{\textit{Study setup}} \\
\texttt{drug\_id\_options}           & Drug\_A, Drug\_B, Drug\_C \\
\texttt{num\_individuals\_range}     & [5, 10] \\
\texttt{num\_peripherals\_range}     & [1, 3] \\
\texttt{solver\_method}              & rk4 \\

\midrule
\multicolumn{2}{l}{\textit{Time grid}} \\
\texttt{time\_start}                 & 0.0 \\
\texttt{time\_stop}                  & 24.0 \\
\texttt{time\_num\_steps}            & 100 \\

\midrule
\multicolumn{2}{l}{\textit{Absorption parameters ($k_a$)}} \\
\texttt{log\_k\_a\_mean\_range}      & [-1, 2] \\
\texttt{log\_k\_a\_std\_range}       & [0.15, 0.45] \\
\texttt{k\_a\_tmag\_range}           & [0.01, 0.1] \\
\texttt{k\_a\_tscl\_range}           & [1, 5] \\

\midrule
\multicolumn{2}{l}{\textit{Elimination parameters ($k_e$)}} \\
\texttt{log\_k\_e\_mean\_range}      & [-5, 0] \\
\texttt{log\_k\_e\_std\_range}       & [0.15, 0.45] \\
\texttt{k\_e\_tmag\_range}           & [0.01, 0.1] \\
\texttt{k\_e\_tscl\_range}           & [1, 5] \\

\midrule
\multicolumn{2}{l}{\textit{Volume parameters ($V$)}} \\
\texttt{log\_V\_mean\_range}         & [1, 7] \\
\texttt{log\_V\_std\_range}          & [0.15, 0.45] \\
\texttt{V\_tmag\_range}              & [0.001, 0.01] \\
\texttt{V\_tscl\_range}              & [1, 5] \\

\midrule
\multicolumn{2}{l}{\textit{Peripheral exchange ($k_{1p}$)}} \\
\texttt{log\_k\_1p\_mean\_range}     & [-4, 0] \\
\texttt{log\_k\_1p\_std\_range}      & [0.15, 0.45] \\
\texttt{k\_1p\_tmag\_range}          & [0.01, 0.1] \\
\texttt{k\_1p\_tscl\_range}          & [1, 5] \\

\midrule
\multicolumn{2}{l}{\textit{Peripheral return ($k_{p1}$)}} \\
\texttt{log\_k\_p1\_mean\_range}     & [-4, -1] \\
\texttt{log\_k\_p1\_std\_range}      & [0.15, 0.45] \\
\texttt{k\_p1\_tmag\_range}          & [0.01, 0.1] \\
\texttt{k\_p1\_tscl\_range}          & [1, 5] \\

\midrule
\multicolumn{2}{l}{\textit{Observation noise}} \\
\texttt{rel\_ruv\_range}             & [0.01, 0.1] \\

\bottomrule
\end{tabular}
\end{table}

\newpage

\section{Operator Encoder-Decoder Transformer}
\label{app:architecture}

% ------------- copy from here -------------
\newcommand{\btau}{\boldsymbol{\tau}}
\newcommand{\bY}{\mathbf{Y}}
\newcommand{\bz}{\boldsymbol{z}}

In the following we describe in detail the model architecture. First we introduce the Continuum operator attention underlying our transformer architecture, then we describe the model's encoder and decoder.

% ---------------------------------------------------------------------
\subsection{Continuum Operator Attention}
\label{app:op_attn}

The standard attention mechanism implicitly treats all key positions as
equally spaced, effectively applying a uniform Riemann-sum approximation to an
underlying integral operator. On an irregular time grid this is a biased estimator.
Following \citet{calvello2024continuum}, we replace uniform averaging with a trapezoidal quadrature correction, turning each attention module into a consistent discretization of a continuum integral operator.

\paragraph{Quadrature weights.}
Given a sorted observation time grid $\tau_1 \le \tau_2 \le \cdots \le \tau_M$ of a subject (either in the context study or the target, depending on the module), define
increments $\Delta\tau_k = \tau_k - \tau_{k-1}$ with $\Delta\tau_1 = 0$,
and the trapezoidal weights
\begin{equation}
    w_k = \tfrac{1}{2}\left(\Delta\tau_k + \Delta\tau_{k+1}\right),
    \quad k = 1, \ldots, M,
    \label{eq:trap_weights}
\end{equation}
setting $\Delta\tau_{M+1} = 0$.
Padding positions are assigned weight zero and the remaining weights are
normalized to unit sum before being passed to the attention modules.

\paragraph{Operator attention.}
Let $\bm{Q} \in \mathbb{R}^{N_q \times d_A}$,
$\bm{K} \in \mathbb{R}^{M \times d_A}$,
$\bm{V} \in \mathbb{R}^{M \times d_A}$ be the query, key, and value matrices
for a single attention head, where $d_A = d / H$ is the per-head dimension, with $H$ the number of  and $d$ is the hidden dimension. The scaled dot-product attention is defined as 
\begin{equation}
{\bm S} = {\bm Q \bm K}^\top / \sqrt{d_A} + \bm{M}
\end{equation}
where $\bm{M} \in \mathbb{R}^{N_q\times M}$ is the attention masking matrix. The operator attention output at query position $i$ is given by:
\begin{equation}
    \Bigl[\mathtt{OpAttn}(\bm{Q}, \bm{K}, \bm{V}; \bm{M}, \bm{\Delta\tau})\Bigr]_i
   \ =\
    \frac{%
        \displaystyle\sum_{k=1}^{M}
        e^{S_{ik}} \cdot w_k \cdot {V}_k
    }{%
        \displaystyle\frac{1}{2}
        \sum_{k=2}^{M}
        \bigl(e^{S_{ik}} + e^{S_{i,k-1}}\bigr)\,\Delta\tau_k
    },
    \label{eq:op_attn}
\end{equation}
where the numerator is a midpoint-rule approximation of the operator 
$\int e^{q(\tau)\cdot k(\tau)}\,v(\tau)\,d\tau$
and the denominator is a trapezoidal approximation of the partition function
$\int e^{q(\tau)\cdot k(\tau)}\,d\tau$.
As the grid spacing vanishes uniformly, Eq.~\eqref{eq:op_attn} converges to
the normalized integral operator of \citet{calvello2024continuum}.
In the limit of a uniform grid with spacing $1/M$, both reduce to standard
softmax attention.

For \emph{operator self-attention},
$(\bm{Q}, \bm{K}, \bm{V})$ are linear projections of the same input observations, and
$\bm{\Delta\tau}$ are the quadrature weights of the observation's time grid. For \emph{operator cross-attention}, $\bm{Q}$ is projected from the query observations,
while $(\bm{K}, \bm{V})$ and $\bm{\Delta\tau}$ are projected from, and computed over,
the key-value observations and their corresponding time grids. Explicitly:
\begin{align}
\mathtt{SelfOpAttn}(\bm{x},\bm{\Delta \tau}, \bm{M}) &= {\tt OpAttn}(\bm{W_Q}\cdot \bm{x}, \bm{W_K}\cdot \bm{x}, \bm{W_V}\cdot \bm{x}\,; \bm{M}, \bm{\Delta\tau})\,,\\
\mathtt{CrossOpAttn}(\bm{x}, \bm{y},\bm{\Delta \tau}, \bm{M}) &= {\tt OpAttn}(\bm{W_Q}\cdot \bm{x}, \bm{W_K}\cdot \bm{y}, \bm{W_V}\cdot \bm{y}\,; \bm{M},   \bm{\Delta\tau})\,,
\end{align}
where $\bm{W_{ Q,K,V}}$ are the corresponding (learnable) linear transformations to the matrices $(\bm{Q}, \bm{K}, \bm{V})$, respectively.

% ---------------------------------------------------------------------
\subsection{Flow-Time Conditioning and Triangular Past--Future Masking}
\label{app:flow_time}

\paragraph{Flow-time embedding.}
The ODE time $t \in [0,1]$ is encoded as a $d$-dimensional sinusoidal Fourier
feature vector with log-spaced frequencies,
\begin{equation}
    t_{\rm emb}
    = \Bigl[
        \sin(t\,\omega_1),\ldots,\sin(t\,\omega_{d/2}),\;
        \cos(t\,\omega_1),\ldots,\cos(t\,\omega_{d/2})
    \Bigr]
    \;\in\; \mathbb{R}^d,
    \label{eq:time_emb}
\end{equation}
where $\omega_k = f_{\max}^{-(k-1)/(d/2-1)}$ for $k = 1,\ldots, d/2$,
giving frequencies log-spaced between 1 and $1/f_{\max}$.
The embedding $t_{\rm emb}$ is added residually to every feature embedding in both encoder
and decoder after each attention block, ensuring the velocity field is a
smooth function of the ODE time at every network layer.
Note that $t$ is the \emph{flow} ODE time and plays no role in the observation
time axis $\tau$; the two are kept strictly separate throughout.

\paragraph{Reference measure and GP posterior.}
\label{app:ref_measure}
As formalised in Eq.~(14) of the main text, the reference measure
$\eta_F(\cdot \mid o, p)$ depends on whether a past prefix is available.
\begin{itemize}
\item In \emph{population synthesis} ($p = 0$), the flow's source is sampled from a zero-mean GP prior $\mathbf{z}_0 \sim \mathrm{GP}(0,\, k)|_{\boldsymbol{\tau}}$ with the standard RBF kernel evaluated at the target's time grid:
\begin{equation}\label{eq:rbf_kernel}
k(\tau,\tau') = \nu^2 \exp \bigl(-\frac{(\tau-\tau')^2}{4\ell^2}\bigr)\,.
\end{equation}
Here, the variance $\nu^2$ and the length scale $\ell$ are model hyperparameters. Since the (normalized) concentrations take values in $\mathbb{R}^+$ we apply a softplus transformation $\mathbf{z}_0 \to {\tt \log(1+\exp(\mathbf{z}_0 ))}$ to the output of the GP that guarantees positive concentrations values along the flow trajectories. 

\item In \emph{forecasting} ($p > 0$), the flow's source $\mathbf{z}_0 = (\mathbf{y}_{\tt P}, \mathbf{x}_{\tt F})$ is sampled from the GP
\emph{posterior} conditioned on the observed past $(\boldsymbol{\tau}_{\tt P}, \mathbf{y}_{\tt P})$:
\begin{equation}
    \mathbf{x}_{\tt F} \sim \mathrm{GP}\bigl(\bm{\mu}_*(\boldsymbol{\tau}_{\tt F}),\;
    \bm{\Sigma}_*(\boldsymbol{\tau}_{\tt F}, \boldsymbol{\tau}_{\tt F})\bigr),
    \label{eq:gp_post}
\end{equation}
where $\bm{\mu}_*$ and $\bm{\Sigma}_*$ are the posterior mean and covariance obtained via GP regression with \eqref{eq:rbf_kernel}. Furthermore, we apply a softplus transformation to the future $\bf x_{\tt F}$, leaving the past unchanged. The posterior already encodes continuity and monotonic behaviour compatible with early-phase PK profiles, providing a warm start for the flow that
substantially reduces the variance of the flow-matching objective compared to
an uninformed prior.
In both cases, past-slot values are fixed to the observed concentrations in
both $\mathbf{z}_0 = (\mathbf{y}_{\tt P}, \mathbf{x}_{\tt F})$ and $\mathbf{z}_1 = (\mathbf{y}_{\tt P}, \mathbf{y}_{\tt F})$
(see Eq.\ (15)), so the interpolation is trivial for past observations.
\end{itemize}

\paragraph{Triangular past--future masking.}
The prefix-dependent mask $\mathbf{M}_p$ in the flow-matching loss Eq.~(16) zeros out the velocity field at all past observations:
\begin{equation}
    v^\theta(t, \mathbf{z}_t, \mathcal{S})
    \;\leftarrow\;
    v^\theta(t, \mathbf{z}_t, \mathcal{S}) \odot \mathbf{M}_p,
    \qquad
    \mathbf{M}_p = \mathbf{1}\!\bigl[\tau > \tau_p\bigr].
    \label{eq:tri_mask}
\end{equation}
This is the discrete implementation of the triangular map structure from
Definition~3.1: no probability mass is transported at past time points, and
their values therefore remain exactly equal to the observed prefix
irrespective of the ODE time $t$.
Combined with the GP-posterior source~\eqref{eq:gp_post}, the flow
learns to move probability mass only in the unobserved future, while the past
acts as a fixed conditioning signal entering through the decoder transformer
~\eqref{app:decoder}.

\subsection{Context Study encoder}

A study $\mathcal{S} = \{\mathcal{D}^1, \ldots, \mathcal{D}^I\}$ contains
$I$ individuals. After normalization (described below), individual $i$ contributes
$C$ time slots (zero-padded where fewer observations exist), yielding a total number of observations $N = I \cdot C$. Each observation in the context study is described by $\mathcal{D}_j^i =({\tau}^i_j,\;{y}^i_j,\;a^i,\;r^i\bigr) \in \mathbb{R}^{4}$, $i\in[I]$, $j\in[C]$, where ${\tau}^i_j$ and ${y}^i_j$ are observation time and concentration, and $(a^i, r^i)$ are the dose amount and administration route. Concentrations are normalized per-study by the maximum concentration observed across all context individuals and time points; times are normalized by the maximum observation time in the study. Both scale factors are estimated from the context set $\mathcal{S}$ and applied consistently to all context and target quantities. The varying number of observation times per individual are tracked by a binary mask $\bm{m}^i\in \{0,1\}^C$ distinguishing true observations from zero-padded entries.

\paragraph{Input embedding}
Each study observation $\mathcal{D}_j^i$ is lifted to the
model dimension via a three-layer MLP $\Phi_{\mathcal S}:\mathbb{R}^4\to \mathbb{R}^d$ interleaved with $\rm GELU$ activations and combined with
a flow-time embedding (Section~\ref{app:flow_time}):
\begin{equation}
    \bm{h}^{i\,(0)}_j
    = \mathtt{LN} \Bigl[
        \Phi_{\mathcal{S}}\!\left(\mathcal{D}_j^i\right)
        + t_{\rm emb}
    \Bigr]
    \;\in\; \mathbb{R}^d,
    \label{eq:enc_embed}
\end{equation}
where $\tt LN(\cdot)$ stands for layer norm.

\paragraph{Encoder}
The study encoder applies $L_E$ stacked operator self-attention blocks $\Psi_\mathcal{S}^{\ell}:\bm{h}^{(\ell-1)}\to\bm{h}^{(\ell)}$, where we denote by ${\bm h}^{(\ell)} \in \mathbb{R}^{N \times d}$ the hidden state after block $\ell\in L$:
\begin{align}
    \bm h &\;=\; \bm{h}^{(\ell-1)}
    + \mathtt{SelfOpAttn}\bigl(\mathtt{LN}(\bm{h}^{(\ell-1)}),\;\bm{\Delta\tau},\;\bm{M}\bigr),
    \label{eq:enc_sa}\\
    \bm{h}^{(\ell)} &\;=\; \bm{h} + \Phi^{\ell}\bigl({\tt LN}(\bm{h})\bigr) + t_{\rm emb}\,\bm{1}^\top_N\,.
    \label{eq:enc_ffn}
\end{align}
Here, $\Phi^{\ell}$ is a two-layer GELU MLP with expansion ratio $4\times$ and $\bm\Delta\tau$ denotes the subject-specific time increments used to form the trapezoidal quadrature weights (Eq.~\eqref{eq:trap_weights}). To make the study-encoding depend explicitly on the functional flow-matching time $t$, we add after each block the time embedding $t_{\rm emb}$ broadcast to each observation in the study. Since subjects in the study are exchangeable but observations between subjects should not mix in the encoder, we use an attention mask $\bm M\in\mathbb{R}^{N\times N}$ that enforces \emph{subject-wise} attention only: we apply a block-diagonal pattern so that tokens from subject \(j\) may attend only to tokens from the same subject \(j\), preventing information flow across subjects within the encoder self-attention. Finally, the output of the encoder yields the hidden study representation:
\begin{equation}
\bm h^{\mathcal S} = {\tt LN}\bigl[\Psi^{L_E-1}_\mathcal{S}(\bm h^{(L_E-1)})\,\circ\,\cdots\circ\Psi^0_\mathcal{S}(\bm h^{(0)})\bigr]\,.
\end{equation}

\subsection{Target Individual Decoder}
\label{app:decoder}
For the observations of a new individual $\mathcal{D}_j = (\tau_j, y_j, a, r)\in \mathbb{R}^4$,
the decoder operates over an irregular grid of $T$ time points formed by concatenating the
observed prefix times $\boldsymbol{\tau}_{\tt P}$ and the future query
times $\boldsymbol{\tau}_{\tt F}$, sorted chronologically.
At flow time $t$, the flow interpolated concentration can be written as
\begin{equation}
    \mathbf{z}_t \;= \;({\bf y}_{\tt P}, {\bf y}_{t,\tt F}) \;= \;\bigl[t\,({\bf y}_{\tt P}, {\bf y}_{\tt F})\; +\; (1-t)\,({\bf y}_{\tt P}, \bf x_{\tt F})\; + \;\sigma_{\min}\,\bm \varepsilon\bigr] \odot \mathbf{M}_p
    \label{eq:zt_def}
\end{equation}
where $\bm\varepsilon \sim \mathcal{N}(0,\bm 1)$ and $\sigma_{\rm min}$ is a small hyperparameter that regularizes the conditional Dirac paths. The irregularity of the observation time grid for target individuals is kept track by a binary mask $\bm{m}\in \{0,1\}^T$ distinguishing true observations from zero-padded entries.

\paragraph{Input embedding.}
Each target individual observation $\mathcal{D}_j \in \mathbb{R}^4$ is embedded by a separate three-layer GELU MLP interleaved with GELU activations and added to the flow-time embedding:
\begin{equation}
    \bm{g}^{(0)}_j
    = \mathtt{LN} \Bigl[
        \Phi_{\mathcal{T}}\!\left(\mathcal{D}_j\right)
        \;+\; t_{\rm emb}
    \Bigr]
    \;\in\; \mathbb{R}^d,
    \label{eq:enc_embed}
\end{equation}
where $\tt LN(\cdot)$ stands for layer norm.

\paragraph{Decoder}
The decoder applies $L_D$ blocks, each containing two attention components with non-causal structure: an intra-individual operator self-attention layer followed by a operator cross-attention layer over the context study $\mathcal{S}$. Writing $\bm{g}^{(\ell)} \in \mathbb{R}^{T \times d}$ for the hidden state after block $\ell\in L_{D}$, the decoder block $\Psi_\mathcal{T}^{\ell}:\bm{g}^{(\ell-1)}\to\bm{g}^{(\ell)}$ is defined by
\begin{align}
    \bm g &\;=\; \bm{g}^{(\ell-1)}
   \; +\; \mathtt{SelfOpAttn}\bigl(\mathtt{LN}(\bm{g}^{(\ell-1)}),\;(\bm{\Delta\tau}_{\tt P}, \bm{\Delta\tau}_{\tt F}),\;\bm{M}_1\bigr),
    \label{eq:enc_sa}\\
    \bm g^\prime &\;=\; \bm{g} \;+\; \Phi_1^{\ell}\bigl({\tt LN}(\bm{g})\bigr)\,,\\
    \bm g^{\prime\prime} &\;=\; \bm{g^{\prime}}
    \;+\; \mathtt{CrossOpAttn}\bigl(\mathtt{LN}(\bm g^\prime)\,,\; \bm{h}^{\mathcal{S}},\;\bm{\Delta\tau},\;\bm{M}_2\bigr)\\
    \bm g^{(\ell)} &\;=\; \bm{g}^{\prime\prime} \;+ \;\Phi_2^{\ell}\bigl({\tt LN}(\bm{g}^{\prime\prime})\bigr)\; + \;t_{\rm emb}\,\bm{1}^\top_T,
\end{align}

where $(\bm\Delta\bm\tau_{\tt P},\bm\Delta\bm\tau_{\tt F})$ and $\bm\Delta\bm\tau$ are the irregular times increments of the (past, future) target individual and study context, respectively, $\Phi_{1,2}$ are 2-layered MLPs and $\bm{h}^\mathcal{S}$ is the study encoder output used for the cross-attention key and values. The attention mask matrices, $\bm M_1\in {\mathbb{R}^{T\times T}}$ and $\bm M_2\in {\mathbb{R}^{T\times N}}$, are defined through their respective binary observation masks, preventing queries to attend to invalid zero-padded entries. This setup allows for information to flow freely between the target's future observations $\bf{y}_{\tt F}$ and the past contextual observations $\bf{y}_{\tt P}$ during self-attention, while subsequently cross-attending to all study observations at all possible times. The output of the target individual decoder reads:
\begin{equation}
\bm g^{\mathcal T} = {\tt LN}\bigl[\Psi^{L_D-1}_\mathcal{T}(\bm g^{(L_D-1)})\,\circ\,\cdots\circ\Psi^0_\mathcal{T}(\bm g^{(0)})\bigr]\,.
\end{equation}

\paragraph{Output head.} Finally, the decoder's output feeds into the head of the architecture parametrized by a 3-layered MLP $\Phi_{\rm head}$:
\begin{equation}
    \bigl[v^\theta_t(\mathbf{z}_t,\mathcal{S})\bigr]_j
    = \mathrm{\Phi}_{\mathrm{head}}\bigl(\bm g^{\mathcal{T}}_j\bigr)\,.
    \label{eq:head}
\end{equation}
This yields the functional velocity field evaluated at all query times.

% ---------------------------------------------------------------------
\subsection{B.7 \quad Hyperparameters and Training Details}
\label{app:hyperparams}

Table~\ref{tab:hyperparams} summarizes the architectural and training
hyperparameters used for the \texttt{PFF} results reported in Section~5.

\begin{table}[h]
\centering
\caption{Hyperparameters of \texttt{PFF}.}
\label{tab:hyperparams}
\begin{tabular}{lc}
\toprule
\textbf{Parameter} & \textbf{Value} \\
\midrule
\multicolumn{2}{l}{\textit{Architecture}} \\
Hidden dimension $d$                        & 256 \\
Encoder depth $L_E$                         & 4   \\
Decoder depth $L_D$                         & 4   \\
Attention heads (encoder / decoder) $H$     & 4 / 4 \\
FFN expansion ratio                         & $4\times$ \\
Dropout                                     & 0.1 \\
Fourier max.\ frequency $f_{\max}$          & 256 \\
\midrule
\multicolumn{2}{l}{\textit{Reference measure (GP regression)}} \\
Kernel variance $\nu^2$                     & $1\!\times\!10^{-4}$ \\
Length scale $\ell$                         & $1.7\!\times\!10^{-3}$ \\
Cholesky Jitter $\epsilon$                           & $1\!\times\!10^{-7}$ \\
Output transform                            & Softplus \\
\midrule
\multicolumn{2}{l}{\textit{Flow matching}} \\
Path noise $\sigma_{\min}$                  & $10^{-4}$ \\
ODE integration steps (inference)           & 100 \\
\midrule
\multicolumn{2}{l}{\textit{Optimisation}} \\
Optimizer                                   & AdamW \\
$(\beta_1,\,\beta_2)$                       & $(0.9,\;0.999)$ \\
Learning rate                               & $1\!\times\!10^{-5}$ \\
Gradient clip (global $\ell_2$ norm)        & 0.5 \\
Batch size                                  & 64 \\
Epochs                                      & 300 \\
Warm-up epochs                              & 5 \\
\bottomrule
\end{tabular}
\end{table}

\end{document}